%% file: main.tex
\documentclass[acmtog]{acmart}

\copyrightyear{2024}
\acmYear{2024}
\setcopyright{acmlicensed}
\acmConference[SA Conference Papers '24]{SIGGRAPH Asia 2024 Conference Papers}{December 3--6, 2024}{Tokyo, Japan}
\acmBooktitle{SIGGRAPH Asia 2024 Conference Papers (SA Conference Papers '24), December 3--6, 2024, Tokyo, Japan}
\acmDOI{10.1145/3680528.3687644}
\acmISBN{979-8-4007-1131-2/24/12}

\acmSubmissionID{701}

\usepackage{graphicx}

\usepackage{booktabs}

\usepackage{multirow}
\usepackage{amsmath,bm}
\usepackage{url}
\usepackage[ruled]{algorithm2e} %

\SetAlFnt{\small}
\SetAlCapFnt{\small}
\SetAlCapNameFnt{\small}
\SetAlCapHSkip{0pt}
\usepackage{subcaption}
\usepackage{float}


\title{DifFRelight: Diffusion-Based Facial Performance Relighting}

\author{Mingming He}
\orcid{0000-0002-9982-7934}
\affiliation{\institution{Netflix Eyeline Studios}
\country{United States of America}}
\email{hmm.lillian@gmail.com}
\authornote{Authors contributed equally to this work.}

\author{Pascal Clausen}
\orcid{0009-0000-2534-9266}
\affiliation{\institution{Netflix Eyeline Studios}
\country{Canada}}
\email{pascal.clausen@scanlinevfx.com}
\authornotemark[1]

\author{Ahmet Levent Taşel}
\orcid{0009-0002-7150-0160}
\affiliation{\institution{Netflix Eyeline Studios}
\country{Canada}}
\email{leventtasel@gmail.com}
\authornotemark[1]

\author{Li Ma}
\orcid{0000-0002-6992-0089}
\affiliation{\institution{Netflix Eyeline Studios}
\country{United States of America}}
\email{lmaag@connect.ust.hk}
\authornotemark[1]

\author{Oliver Pilarski}
\orcid{0009-0005-7803-2314}
\affiliation{\institution{Netflix Eyeline Studios}
\country{Germany}}
\email{oliver.pilarski@scanlinevfx.com}
\authornotemark[1]
\authornote{Corresponding author.}

\author{Wenqi Xian}
\orcid{0009-0007-4678-6458}
\affiliation{\institution{Netflix Eyeline Studios}
\country{United States of America}}
\email{wenqixian3@gmail.com}

\author{Laszlo Rikker}
\orcid{0009-0005-2469-3440}
\affiliation{\institution{Netflix Eyeline Studios}
\country{United Kingdom}}
\email{laszlo.rikker@scanlinevfx.com}

\author{Xueming Yu}
\orcid{0009-0009-8189-6024}
\affiliation{\institution{Netflix Eyeline Studios}
\country{United States of America}}
\email{xueming.yu@scanlinevfx.com}

\author{Ryan Burgert}
\orcid{0009-0008-5947-2076}
\affiliation{\institution{Netflix Eyeline Studios}
\country{United States of America}}
\email{ryancentralorg@gmail.com}

\author{Ning Yu}
\orcid{0009-0004-6865-1325}
\affiliation{\institution{Netflix Eyeline Studios}
\country{United States of America}}
\email{ningyu.hust@gmail.com}

\author{Paul Debevec}
\orcid{0000-0001-7381-2323}
\affiliation{\institution{Netflix Eyeline Studios}
\country{United States of America}}
\email{debevec@gmail.com}

\begin{document}

\begin{abstract}
We present a novel framework for free-viewpoint facial performance relighting using diffusion-based image-to-image translation. Leveraging a subject-specific dataset containing diverse facial expressions captured under various lighting conditions, including flat-lit and one-light-at-a-time (OLAT) scenarios, we train a diffusion model for precise lighting control, enabling high-fidelity relit facial images from flat-lit inputs. Our framework includes spatially-aligned conditioning of flat-lit captures and random noise, along with integrated lighting information for global control, utilizing prior knowledge from the pre-trained Stable Diffusion model. This model is then applied to dynamic facial performances captured in a consistent flat-lit environment and reconstructed for novel-view synthesis using a scalable dynamic 3D Gaussian Splatting method to maintain quality and consistency in the relit results. In addition, we introduce unified lighting control by integrating a novel area lighting representation with directional lighting, allowing for joint adjustments in light size and direction. We also enable high dynamic range imaging (HDRI) composition using multiple directional lights to produce dynamic sequences under complex lighting conditions. Our evaluations demonstrate the model's efficiency in achieving precise lighting control and generalizing across various facial expressions while preserving detailed features such as skin texture and hair. The model accurately reproduces complex lighting effects like eye reflections, subsurface scattering, self-shadowing, and translucency, advancing photorealism within our framework.
\end{abstract}


\begin{teaserfigure}
  \includegraphics[width=\textwidth]{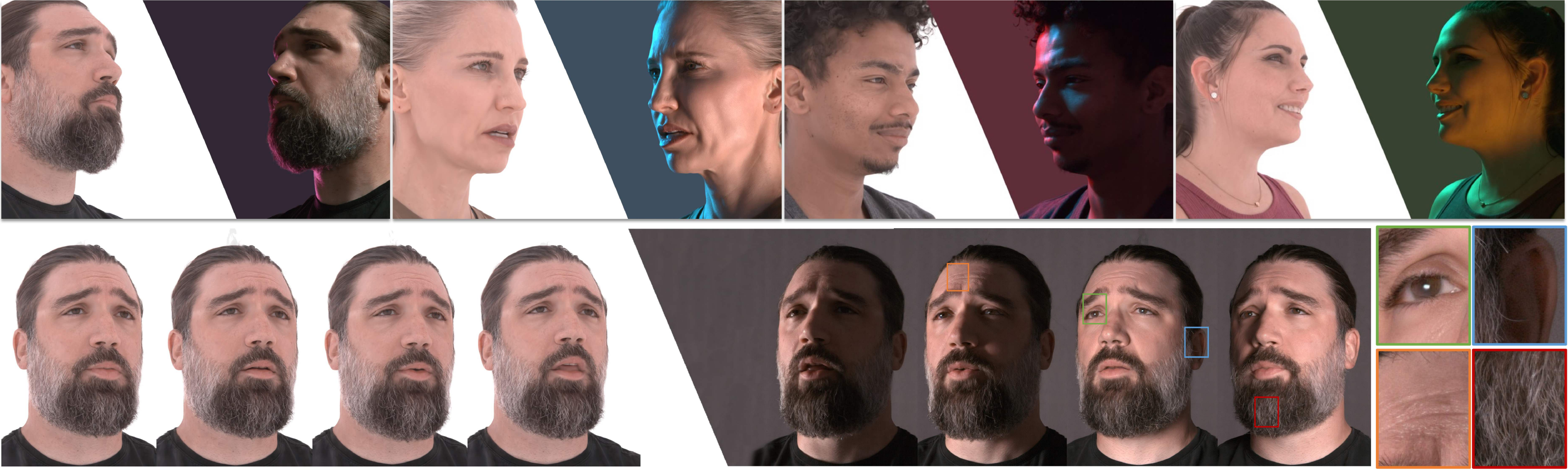}
  \vspace*{-6mm}
  \caption{\textbf{Flat-lit input and relit results of our diffusion-based relighting method.} Top row: four flat-lit subjects transformed to colorful multi-directional lighting. The flat-lit images are novel views generated by our dynamic 3D Gaussian Splatting extension from multi-view video. Bottom row: selected frames from a dynamic flat-lit sequence alongside their relit counterparts with an animated lighting direction. Facial details show our method's high-fidelity relighting effects, including eye reflections, translucency, and self-shadowing.}
  \Description{Relighting of flat-lit captured talents under novel lightings on novel facial expressions and novel views.}
  \label{fig:teaser}
\end{teaserfigure}

\maketitle

\input{intro}

\input{related}

\input{method}

\input{experiments}

\input{limitation}

\input{conclusion}

\begin{acks}
We thank Stephan Trojansky and Jeffrey Shapiro for initiation and ongoing support; Gabriel Dedic for invaluable help on imaging; Jon Milward, Samuel Price for stage operation; Connie Siu, Amir Shachar, Kevin Williams, Marc Thorineaux, Alex Chun for stage coordination; Rafe Sacks, Scott Wilson for data preprocessing; Sunny Koya, Daniel Heckenberg, Pablo Delgado, Elliot Chow for technical support; the software department for stage software maintenance; Brian Tong for video editing, and our performers: Kevin Williams, Mary Carr, Zorianna Kit, and Naz Lang.
\end{acks}

\input{full_results_figures_2pages}

\clearpage
\bibliographystyle{ACM-Reference-Format}
\bibliography{main}

\clearpage
\appendix
\include{appendix}

\end{document}

%% file: intro.tex
\section{Introduction}

Today, photorealistic digital humans often play starring roles alongside live-action actors, and appear with increasing fidelity in video games and virtual reality.  However, constructing photorealistic digital human models is difficult and expensive, and animating them realistically -- even with performance-capture data -- is still challenging and prone to error.

Volumetric performance capture (volcap) systems use an array of inward-pointing cameras to record dynamic human performances in three dimensions, avoiding the need for traditional 3D character modeling, rigging, and animation. Volcap captures all the nuances of a performance, allowing it to be rendered into any 3D scene from any angle. However, most capture techniques record the subject under a single flat lighting condition, which hampers the achievement of cinematic lighting effects and complicates the seamless integration of characters into new scenes. To address this, some advanced systems \cite{Guo:therelightables:2019} use time-multiplexed lighting to record performances under multiple lighting conditions alternating each frame. This approach enables data-driven relighting, sometimes enhanced by machine learning inference \cite{meka2020_deep_relightable_textures,zhang_neural_light_transport_2021}.

In this work, we achieve high-quality relighting of volumetrically captured facial performances as shown in Fig.~\ref{fig:teaser}. This is accomplished by fine-tuning a diffusion-based image-to-image translation model using subject-specific relighting examples. 
In addition to capturing the performance under flat lighting, we record each subject under a dense array of lighting directions across various facial expressions. We train a diffusion model for precise lighting control, enabling the generation of high-fidelity relit facial images from flat-lit inputs. This process involves spatially-aligned conditioning of flat-lit captures and random noise, integrated with lighting information for global control, leveraging prior knowledge from the pretrained stable diffusion model. Additionally, we use dynamic 3D Gaussian Splatting (3DGS) to extrapolate the performance to novel viewpoints.

Our work includes the following contributions:
\begin{enumerate}
    \item A novel framework for facial performance relighting that trains on a subject-specific paired dataset and generalizes to novel lighting conditions, enabling relighting from free viewpoints and unseen facial expressions.
    \item A diffusion-based relighting model that spatially conditions the flat-lit input image, utilizing lighting information as global control to generate high-quality relit results.
    \item A scalable dynamic 3DGS technique for reconstructing long sequences, ensuring temporal consistency in flat-lit inputs for coherent inference by the relighting model.
    \item A unified lighting control that combines a new area lighting representation with directional lighting, offering versatile lighting controls as well as enabling the composition of complex environment lighting.  
\end{enumerate}

%% file: related.tex
\begin{figure*}[t]
    \centering
    \includegraphics[width=1.\linewidth]{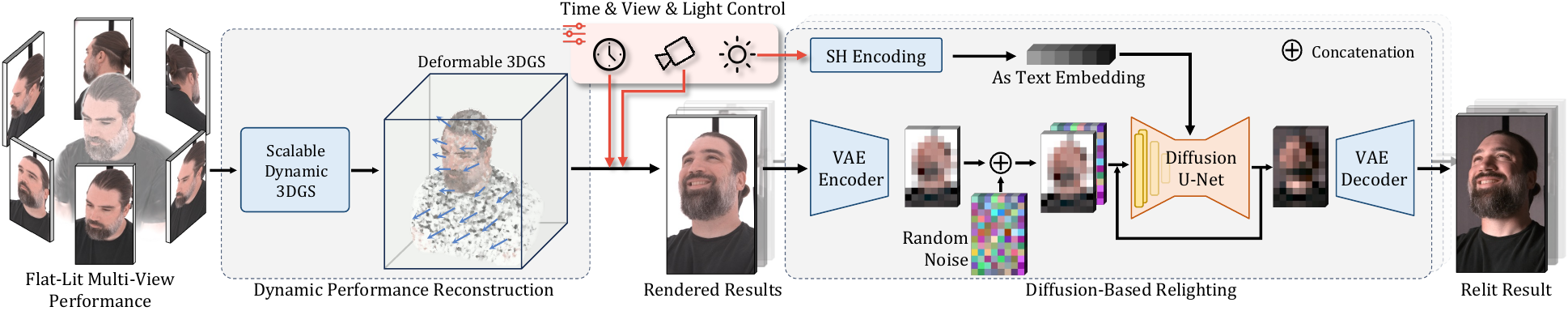}
    \vspace{-6mm}
    \caption{\textbf{The overview of the proposed relighting pipeline including dynamic performance reconstruction and diffusion-based relighting.} Starting with multi-view performance data of a subject in a neutral environment, we train a deformable 3DGS to create novel-view renderings of the dynamic sequence. These serve as inputs for a diffusion-based relighting model, trained on paired data to translate flat-lit input images to relit results based on specified lighting. Here, we show the inference step of the diffusion model, where the latent representation of the flat-lit image is concatenated with random noise as input for the diffusion U-Net. Lighting information, encoded as SH encoding together with the text embedding, regulates the diffusion process.}
    \label{fig:method_overview}
    \vspace{-2mm}
\end{figure*}

\section{Related Work}

\subsection{Non-Diffusion-Based Relighting Methods}

\textit{Parametric reflectance modeling} measures a subject's geometry and reflectance properties from different views and lighting directions but struggles with complex materials and dynamic subjects~\cite{Sato:1997:Reflectance,Goesele:2004:DISCO,weyrich2006_analysis_of_human_faces}. These techniques struggle with complex materials, such as hair, and are difficult to apply to dynamic subjects since observations under many lighting conditions are required.

\textit{Image-based relighting} photographs the subject under various lighting conditions to create new lighting scenarios, which is effective but costly and challenging for dynamic subjects~\cite{nimeroff1995efficient,debevec2000_acquiring_reflectance_field,Einarsson:2006:RHL,peers2007_postproduction_facial}.

\textit{Intrinsic image relighting} decomposes an image into intrinsic components and recomposes it with modified lighting~\cite{wang2008face,barron2014shape,shu2017neural,sengupta2018sfsnet,le2019illumination,tewari2021monocular,zhou2019deep,sun2019single,tewari2020pie,tewari2021photoapp,mei2024holo,shih2014style,luan2017deep,shu2017portrait,li2018closed}. Despite advancements, these methods face challenges with complex shading effects and detail preservation. Recently, \citet{kim2024switchlightcodesignphysicsdrivenarchitecture} propose a physically based rendering model for computing diffuse and specular reflectance, inspired by \cite{pandey2021_total_relighting}. They can generalize to novel subjects but lack detail for skin and hair reflectance, although refining with a neural renderer. \citet{rao2022vorf} develop a model with a reflectance network trained on OLAT images to learn a light-conditioned volumetric reflectance field from a single photo, but it is limited to static faces and does not address dynamic sequences.

\textit{Neural relighting} approaches either use Neural Radiance Fields (NeRFs) to reconstruct material properties and lighting effects in 3D but face challenges in novel poses and detail preservation~\cite{mildenhall2020_nerf,wang2023_neus,boss2021_nerd,zhang2021_nerfactor,liu2023_nero,li2023_relitneulf,zeng2023_relighting_nerf_shadow,sarkar2023_litnerf,srinivasan2021_nerv,yao2022_neilf,zhang2023_neilfpp,xu2023_renerf}, or introduce 3D Gaussian Splatting (3DGS) for relighting because of the fine detail reconstruction~\cite{kerbl2023_3d_gs,gao2023_relightable_3d_gaussians}. Recent research on relightable head avatars uses OLAT data and various geometry representations, such as volumetric primitives~\cite{yang2023practicalcapturehighfidelityrelightable}, neural fields~\cite{xu2023artistfriendlyrelightableanimatableneural}, and 3DGS~\cite{saito2023_relightable_gs_codec_avatars}, to enhance the rendering quality of facial details. The subject-specific model~\cite{saito2023_relightable_gs_codec_avatars} achieves promising facial performance relighting by combining 3D Gaussians with a relightable appearance model based on learnable radiance transfer.
However, it necessitates a significantly denser capture settings for expressions and requires precise track of facial mesh and gaze motion and explicit modeling of face and eyes, making this method difficult to generalize.

\vspace{-1mm}
\subsection{Diffusion-Based Relighting Methods}

Diffusion models ~\cite{song2020score,song2020denoising,rombach2022high,karras2022elucidating,preechakul2022diffusion,zhang2023adding,wang2023exploiting,ke2023repurposing} have demonstrated remarkable capabilities in generating high-quality images by sampling from a learned distribution of images, particularly when conditioned on spatial control via image-to-image translation~\cite{meng2021sdedit,parmar2023zero,brooks2023instructpix2pix,wang2023exploiting,ke2023repurposing}.

Recently, researchers have leveraged conditional diffusion models for relighting~\cite{ding2023diffusionrig,ponglertnapakorn2023difareli,zeng2024dilightnet}. These models leverage their ability to learn complex light interactions, drawing on a generalizable generative prior that has been pretrained on a large dataset of images under various lighting conditions. Additionally, they benefit from their controllability, which is fine-tuned using flat-relit image pairs.
DiffusionRig~\cite{ding2023diffusionrig} proposes to manipulate lighting and beyond by learning a mapping from computer-generated imagery (CGI) to real images using surface normals, albedo, and a diffuse shaded 3D morphable model. DiFaReli~\cite{ponglertnapakorn2023difareli} uses existing estimators for lighting, a 3D morphable model, the subject's identity, camera parameters, and a foreground mask, and trains a conditional diffusion network that takes a diffuse rendered model under novel lighting. Similarly, DiLightNet~\cite{zeng2024dilightnet} adopts a comparable approach, offering multiple radiance hints (diffuse and specular) without requiring a real-world captured input photograph, but is not trained on human or face datasets. Unlike one-shot methods, our approach uses multi-view, multi-pose data captured varying and controlled lighting conditions, allowing us to focus on high-fidelity identity-preserving reconstruction. We aim to generalize to arbitrary views, relighting conditions, and unseen poses, and prioritize preserving the subject identity.

%% file: method.tex
\section{Method}

Fig.~\ref{fig:method_overview} illustrates our dynamic facial performance relighting pipeline with two main components: dynamic 3D performance reconstruction and diffusion-based relighting. Our approach begins with the acquisition of multi-view performance sequences in a flat-lit environment. Utilizing our scalable approach of dynamic 3D Gaussian Splatting (3DGS), we first reconstruct deformable 3D Gaussians to render the sequence from novel perspectives. Then we employ the diffusion-based relighting model to generate new lighting for the rendered image sequence, based on a specified lighting direction. This model has been trained with subject-specific paired data of flat-lit and OLAT images captured using a customized LED-panel stage (Sec.~\ref{sec:data_capture}), incorporating a new design of feeding flat-lit images and lighting directions as spatial and global conditions into the pretrained latent diffusion model (Sec.~\ref{sec:diff_relight}). To maintain temporal consistency in the dynamic relighting, we introduce a partitioning and combining scheme to train temporally-coherent 3DGS for lengthy sequences and also apply temporal blending on the dynamic relit results (Sec.~\ref{sec:dynamic_gs}). Additionally, we introduce a new variable-sized area light representation for enhanced control over different lighting types for various lighting compositions (Sec.~\ref{sec:light_control}).

\subsection{Facial Data Capture}
\label{sec:data_capture}

\begin{figure}[t]
\setlength{\tabcolsep}{0\linewidth}
    \centering
    \includegraphics[width=1.0\linewidth]{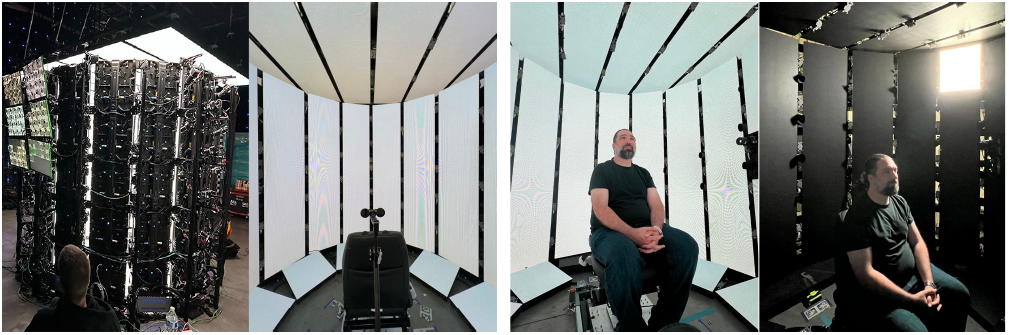}  \\ 
       \begin{small}
       (a) Capture Stage (Inside \& Outside \textit{w/} and \textit{w/o} Subject) \\
        \vspace{1mm}
        \end{small}
    \includegraphics[width=1.0\linewidth]{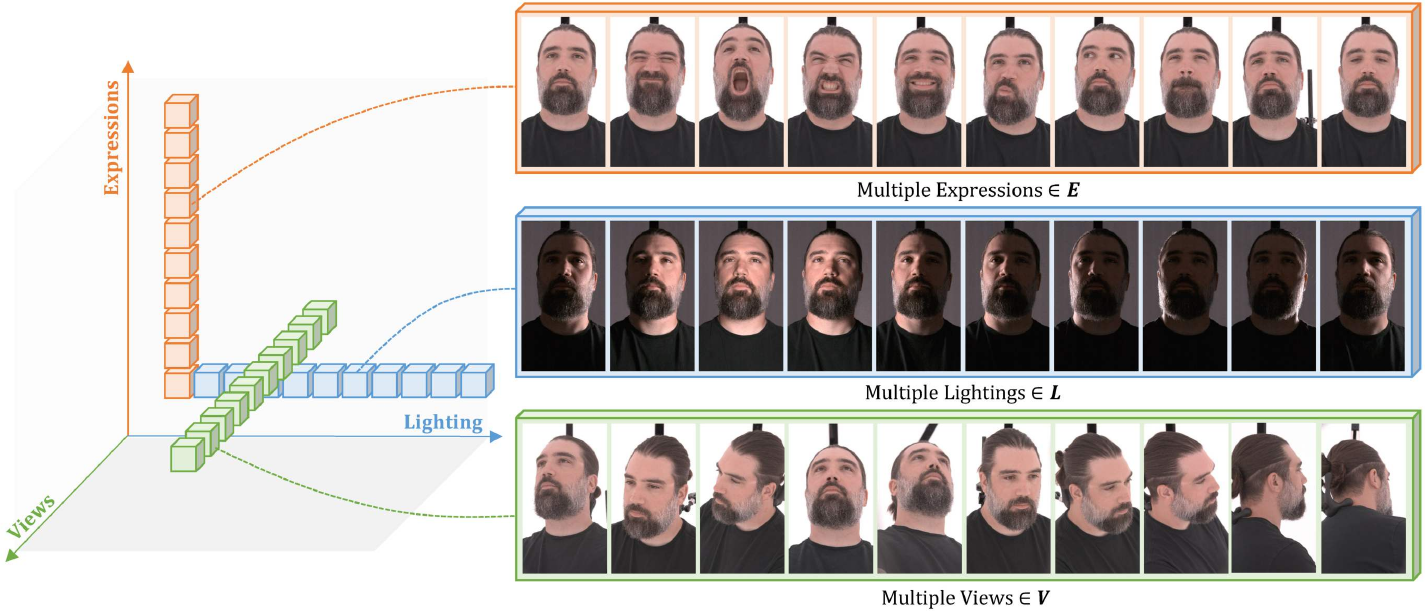} \\
       \begin{small}
       (b) Visualization of Captured Data \\
       \end{small}
       \vspace{-2mm}
    \caption{\textbf{Capture stage and data.} (a) outside and inside of the LED panel volcap stage based on the LED panels, with and without the subject. (b) examples of our collected training data including variation in expression ($\mathbf{E}$), lighting ($\mathbf{L}$), and viewpoint ($\mathbf{V}$). }
    \label{fig:method_facerig}
    \vspace{-3mm}
\end{figure}

\paragraph{Capture Setup} 
To supervise the diffusion-based relighting model and facilitate learning the translation of facial reflectance from flat lighting to arbitrary lighting, we require paired training data where lighting is the only changing variable. We use a volumetric capture stage consisting of a multi-view array camera array placed within a capped cylinder of LED panels as in Fig.~\ref{fig:method_facerig} (a). By turning on different LED panels, we capture the same subject under varying lighting conditions. Furthermore, customizing the display patterns of LED panels enables us to record the subject under various backgrounds and image-based lighting environments. We constructed this stage as a cylindrical capture rig using off-the-shelf ROE BP2v2 LED panels and $75$ synchronized 4K Z-CAM e2 cinema cameras. The rig comprises a $16\times5$ array of $50cm \times 50cm$ panels (16 columns, each consisting of 5 panels), forming a capture volume approximately $250$ cm tall and $276$ cm in diameter. The ceiling is covered with a $6 \times 5$ array of panels, and an additional $10$ panels on the floor provide additional lighting angles. The cameras peer in through $5$cm gaps in the walls and ceiling.

\paragraph{Lighting setup} 
Following~\citet{LeGendre_olat}, we capture the reflectance field as OLAT images using each panel in sequence as a single light source at $24$ frames per second in sync with the camera array. 
As~\citet{debevec2022hdrlightingdilationdynamic} demonstrate, LED panels are effective for image-based lighting on actors, so we use them to simulate OLATs, with one light (=OL) equivalent to one LED panel. Each panel approximates a 20-degree cone from the stage center, producing less aliased shadows than point lights while still capturing eye highlights and sharp specular reflections. We minimize the panel size while maintaining sufficient illumination and try to ensure uniformity in size and shape. Despite the large angular coverage of each panel, in practice, our OLAT-based framework works reasonably well when generalizing to environment lighting, as shown in Fig.~\ref{fig:novellights_hdrs} and Fig.~\ref{fig:real_world_hdri}.
We also capture flat-lit lighting to have paired data for training the diffusion-based relighting model.

\paragraph{Data collection} 
We collected OLAT sequences for four individuals, each demonstrating $30$ facial expressions and head poses. Each OLAT sequence has $123$ lighting conditions with $13$ interspersed flat-lit tracking frames. This same flat-lit lighting is used to capture the dynamic facial performance. Additionally, we recorded one OLAT sequence of a clean plate without a subject to facilitate background removal in 3DGS reconstruction. To compensate for the individual's slight movements during one OLAT sequence, we conduct optical flow alignment in image space with respect to the flat-lit frames for each view, following~\citet{meka2019deep}. When training the relighting model, we convert linear images to sRGB color space for better compatibility with the pretrained model weights.

\subsection{Diffusion-Based Relighting}
\label{sec:diff_relight}
After data acquisition and preprocessing, we obtain a paired dataset $\{\textbf{I}_{FlatLit}, \textbf{I}_{OLAT}, \textbf{d}\}$, where $\textbf{I}_{FlatLit}$ represents the flat-lit image, and $\textbf{I}_{OLAT}$ denotes the OLAT image under light direction $\textbf{d}$. $\textbf{d}$ is defined as a directional vector pointing from the stage center to the center of each panel in camera space. Our objective is to train a personalized model capable of relighting the flat-lit image of the same subject under novel views, novel lightings, and novel expressions/poses. Given the limited set of training pairs we have captured, a strong prior is essential for the model to generalize to unseen conditions. Rather than relying on physically-based lighting models, which often fail to accurately simulate complex shading effects such as anisotropic materials and subsurface scattering, we employ a purely data-driven prior to ensure the photorealism of our output. We use a diffusion-based approach that leverages the prior knowledge encapsulated in the pre-trained Stable Diffusion model \cite{rombach2022high}.

We formulate our problem as an image-to-image translation that transfers a flat-lit image $\textbf{I}_{FlatLit}$ to the corresponding OLAT image $\textbf{I}_{OLAT}$ under the condition $\textbf{d}$. 
Inspired by recent monocular depth estimation works \cite{ke2023repurposing}, we adapt Stable Diffusion's architecture to fine-tune it with our paired data conditioned on lighting information, achieving robust performance in relighting tasks.

The Stable Diffusion model \cite{rombach2022high} is originally designed as a latent diffusion model for text-to-image generation. It comprises a Variational Autoencoder (VAE) that encodes an image $\mathbf{I}$ into the latent space $\mathbf{z} = \mathcal{E}(\mathbf{I})$ and decodes it back to $\mathbf{I} \approx \mathcal{D}(\mathbf{z})$. Additionally, it includes a diffusion U-Net $\hat{\epsilon}(\mathbf{z}^{(t)}; \mathbf{s}, t)$, which predicts the noise of a partially denoised latent $\mathbf{z}^{(t)}$ conditioned on some text embedding $\mathbf{s}$ and the diffusion timestep $t$. By iteratively removing noise from random noise, we obtain a clean image latent $\mathbf{z}^{(0)}$, which can then be decoded back into an image $\mathcal{D}(\mathbf{z}^{(0)})$.

To transform the image generation framework into an image-to-image translation model and more effectively leverage the spatial information in the flat-lit image, we modify the U-Net by concatenating the latent of the flat-lit image with the random noise map. This necessitates a minor change in the network structure: specifically, doubling the input channel count of the first convolutional layer. These minimal adjustments enable us to maximize the retention of the pre-trained weights from Stable Diffusion. Additionally, concatenating the latents with random noise improves the alignment of the spatial structure between the relit result and the input.

In addition to the flat-lit image, which offers spatial cues to the resulting output, the lighting information acts as a crucial global control signal that influences the entire image. We encode light direction $\mathbf{d}$ into higher dimensions and replace text embeddings by introducing it into the U-Net through cross-attention in diffusion models.
To increase the frequency band of $\mathbf{d}$ and enhance the conditioning, we use Spherical Harmonics (SH) to encode $\mathbf{d}$, similar to techniques used by~\citet{tancik2023nerfstudio}.
We also zero-pad the SH encoding to match the length of the text embedding: 
\begin{equation}
    \mathbf{s}_d = \mathbf{0} \oplus \mathcal{Y}(\mathbf{d}),
\end{equation}
where $\oplus$ indicates concatenation, and $\mathcal{Y}$ indicates the SH encoding. 
Initially, this may appear unusual as the SH encoding is nothing like a text embedding, which is predicted by some text encoder \cite{clip}. However, we find that the fine-tuning is adequate for filling the domain gap between the text embedding and SH encoding. We set the SH degree to 3 for all model training, as it showed the best performance in our experiments (see Sec.~5.1 in the supplemental material).

At training time, we freeze the VAE encoder and decoder, as both the flat-lit and OLAT-lit images lie within the manifold of the VAE's latent space. We randomly select training pairs from the dataset and fine-tune the diffusion U-Net using the denoising diffusion objective:

\begin{equation}
    \mathcal{L}_\text{diffusion} = \|\hat{\epsilon}\left(\mathbf{z}_{OLAT}^{(t)} \oplus \mathcal{E}(\mathbf{I}_{FlatLit}); \mathbf{s}_d, t\right) - \epsilon\|_2^2,
\end{equation}
where $\mathbf{z}_{OLAT}^{(t)}$ is a noisy ground truth latent at diffusion time step $t$, combining ground truth latent $\mathcal{E}(\mathbf{I}_{OLAT})$ and random noise map $\epsilon$:
\begin{equation}
    \mathbf{z}_{OLAT}^{(t)} = \sqrt{\alpha_t} \mathcal{E}(\mathbf{I}_{OLAT}) + \sqrt{1 -\alpha_t} \epsilon \text{.}
\end{equation}
Here $\alpha_t$ is a predefined value that schedules the diffusion process. We find that using naive Gaussian noise results in substantial color shift. This is likely due to the inductive bias that the network struggles to predict very dark images using zero-meaned Gaussian noise \cite{offsetnoise} for $\epsilon$.  Pyramid noise, as introduced in \cite{multiresnoise}, greatly improves the color consistency between the prediction and the ground truth. 

We fine-tune one personalized model for each subject using the captured data. During inference, we are able to relight a flat-lit image of the same subject under arbitrary combinations of facial expression, camera pose, and light direction.

\subsection{Dynamic Performance Relighting}
\label{sec:dynamic_gs}

\begin{figure}[t]
    \centering
    \includegraphics[width=1.\linewidth]{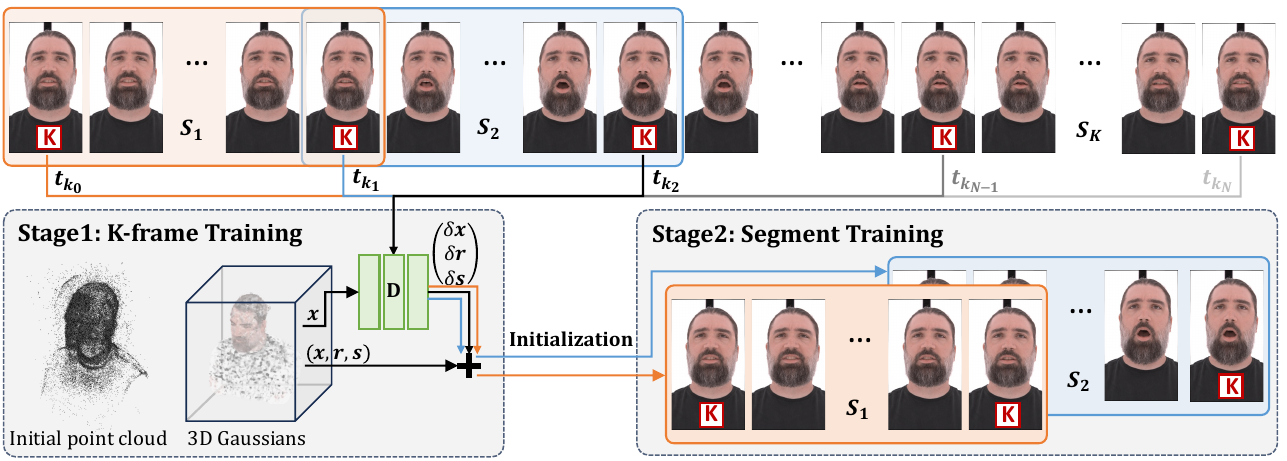}
    \vspace{-5mm}
    \caption{\textbf{Two-stage training of the scalable dynamic 3DGS.} We first sample $K$ frames to partition the long sequence into segments ${S_1, S_2, ...S_k}$. At \textit{Stage 1}, we train the deformable 3DGS on the $K$ frames only to generate the initialization for the training of each segment at \textit{Stage 2}. Then we train a deformable 3DGS for each segment but conditioned on the initialization.}
    \label{fig:method_dgs}
    \vspace{-3mm}
\end{figure}

To ensure temporally consistent relit results when applying the pre-trained diffusion-based relighting model to flat-lit facial performances, our initial goal is to minimize the domain gap between training and testing by generating photorealistic novel-view renderings. Additionally, maintaining temporal consistency in the flat-lit inputs enhances the diffusion model's performance under consistent conditions. In line with these requirements, we utilize 3D Gaussians~\cite{kerbl2023_3d_gs} as the geometry representation for capturing fine details in real-time rendering. 3D Gaussian Splatting (3DGS) can be extended to the temporal domain ~\cite{luiten2023_dynamic_gs,wu2023_4d_gs,jung2023_deformable_gs,li2023spacetime}. Deformable 3DGS~\cite{jung2023_deformable_gs} learns a deformation field for each frame within the canonical space of 3D Gaussians that is shared across all frames. The deformation field takes the positions of the 3D Gaussians $\textbf{x}$ and the current time $t$ as inputs, outputting the offset for position $\textbf{x}$, rotation $\textbf{r}$, and scaling $\textbf{s}$ as $\delta_t \textbf{x}$, $\delta_t \textbf{r}$, and $\delta_t \textbf{s}$. While preserving details in short videos, this method faces challenges in accurately representing motion and maintaining intricate details in long sequences due to its use of globally shared Gaussians, which struggle with the complexities of extended sequences.

We introduce a novel scalable method based on the deformable 3DGS~\cite{jung2023_deformable_gs} to optimize 3D Gaussians for lengthy performance sequences, thereby generating temporally consistent reconstruction. Instead of training all frames together with one shared set of Gaussians, we partition it into small segments with an equal number of frames, allowing varying Gaussians across segments. To minimize temporal inconsistency at the transition frame between segments and to preserve a similar level of reconstruction details across segments, we design a two-stage training strategy. As shown in~Fig.\ref{fig:method_dgs}, in \textit{Stage 1}, we evenly distribute $K$ keyframes in the entire sequence, and then train a deformable Gaussian splatting model based on the keyframes only, generating the pretrained $K$-frame model. In \textit{Stage 2}, using the $K$ frames as transition points, we partition the entire sequence into $K-1$ segments, each containing two keyframes (one at the beginning and the other at the end). Inside every segment, we initialize the 3D Gaussians and the deformation network using the pretrained $K$-frame model based on the time of its first frame in the global time range. The geometry information of 3D Gaussians (such as positions $\textbf{x}$, rotations $\textbf{r}$ and scalings $\textbf{s}$) is initialized as the pretained $K$-frame Gaussians transformed by the deformation offset of the first frame $t_{k_0}$. In this way, we ensure the initial states of 3D Gaussians in different segments are temporally consistent with similar level of details. To further enhance the consistency, we fix the initial Gaussians and train the deformation network only for the warm-up iterations (we use $3000$ iterations as default), with the regularization term defined as $L_2$ loss on the deformation offset of keyframes $k_0$ and $k_1$:
\begin{equation}
\begin{split}
    \mathcal{L}_{reg} & = \|\delta'_{t_{k_0}} - \delta_{t_{k_0}} \|_2 + \| \delta'_{t_{k_1}} - \delta_{t_{k_1}} \|_2,
\end{split}
\end{equation}
where $\delta_{t_{k_0}}$ and $\delta_{t_{k_1}}$ are initial deformation offset of $t_{k_0}$ and $t_{k_1}$ while $\delta'_{t_{k_0}}$ and $\delta'_{t_{k_1}}$ indicates the learned offset with respect to $\textbf{x}$, $\textbf{r}$ and $\textbf{s}$. 

This warm-up training allows the Gaussians to freely deform to reconstruct motions while being restricted to the deformation of keyframes at transition points. After this stage, most Gaussians are in the correct position, and we relax the constraints to allow Gaussians to clone, split, and prune for detailed reconstruction for a number of iterations (defaulting to $10000$ iterations). After the densification, we fix the number of Gaussians and jointly optimize the deformation network and the Gaussians. The ablation of the two-stage training method is included in the supplemental material.

Although we ensure temporal consistency in the reconstructed flat-lit sequence used as input for our relighting model, the image-based diffusion model may still produce temporally inconsistent high frequencies. To address this, we adopt the method from \cite{jamrivska2019stylizing}, applying a temporal blending approach to interpolate relit results between keyframes. We use a small step size of $4$ to sample keyframes and preserve details and lighting accuracy.

\subsection{Unified Lighting Control}
\label{sec:light_control}

To provide the users with flexible lighting control, we propose a novel area lighting representation, which includes both lighting direction and a variable light size. This is integrated with our directional lighting into a unified lighting control to guide the diffusion-based relighting model. Based on the single-lighting inference, we can enable HDRI environment lighting reconstruction by compositing multiple single-lighting inferences.

\paragraph{Area-light training}
To infer lights of various sizes -- from point lighting to diffuse -- we modify our training in two ways. First, we multiply the light direction by a light size factor $1-a$, with $a$ ranging from $0$ (indicating a single OLAT) to $1$ (representing the flat-lit), which represents the spread of the light and thus its sharpness. Second, we construct our ground truth target by combining OLAT images to simulate isotropic Spherical Gaussian (SG) illumination. Then the light size factor is linked to the Spherical Gaussian's sharpness by the formula $\lambda=\frac{-cos(\theta)}{(cos(\theta)^2-1)}$ with $\theta=a \cdot (\theta_{max} - \theta_{min})+\theta_{min}$, $\frac{\pi}{2}$ being the inflection point of an SG and $\theta_{min}=1 \cdot \frac{\pi}{180}$, $\theta_{max}=89 \cdot \frac{\pi}{180}$.

\paragraph{HDRI reconstruction}
To relight a performance under an HDRI environment using our model conditioned on a single lighting, we first map an HDRI latitude-longitude (lat-long) representation to the OLATs, with each OLAT’s weighting coefficient being an average from its corresponding area on the HDRI lat-long. During inference, we generate an image for each OLAT direction and compute a weighted sum of these OLATs in linear space using their fitted coefficients. Additionally, the HDRI map can be reconstructed using our model conditioned on both light direction and size by first fitting Spherical Gaussians onto the HDRI lat-long representation and then inferring for each light direction and size. Similarly to the OLATs, we compute a weighted sum of the predictions using the fitted coefficients of these Spherical Gaussians. If the HDRI is animated, we adjust the conditioned directions of the OLATs or Spherical Gaussians to match the rotation of the HDRI.

%% file: experiments.tex
\section{Experiments}
We have conducted multiple experiments to evaluate the performance of our diffusion-based relighting model, particularly its generalization across various dimensions including novel lightings, expressions, and views. We compare our model with two baselines constructed using different network structures from the related methods. Additionally, we perform ablation studies to verify various technical designs within our system and conduct a simple extension to assess model generalization to novel subjects. For more results on dynamic performance relighting and detailed experiments, please refer to the accompanying video and supplemental material.

\subsection{Implementation Details}
We capture 30 expressions ($30E$), 123 OLATs ($123L$), and 75 views ($75V$) for each of the 4 subjects and downsample 4K captures to $1080 \times 1920$. For training, we use the Cartesian product of \(27E \times 115L \times 69V\), with the rest reserved for validation. Capturing all OLAT data for each subject takes a few minutes. For inference, we record 10-15 seconds of performances in a multi-view capture stage under flat lighting.
We implement our diffusion-based relighting system using Diffusers \cite{HuggingFace_Diffusers} based on Stable Diffusion v2.1.
Our custom models are trained on 8 NVIDIA A100 GPUs with 40GB memory each, for 100K iterations with a batch size of 8. We use an Adam optimizer with a learning rate of $3 \cdot 10^{-5}$. We use a DDPM scheduler with 1000 steps for training and a DDIM scheduler with 30 steps for inference which takes about 5-6 seconds per frame on a single A100 GPU.
Our dynamic 3DGS is also trained in full HD resolution. Given a long sequence, we divide it into segments based on the number of keyframes, typically including $20$ frames per segment. We then train the deformable 3DGS for 40K iterations for each segment.

\subsection{Results}
\begin{figure}
    \centering
    \vspace{-1mm}
    \includegraphics[width=0.99\linewidth]{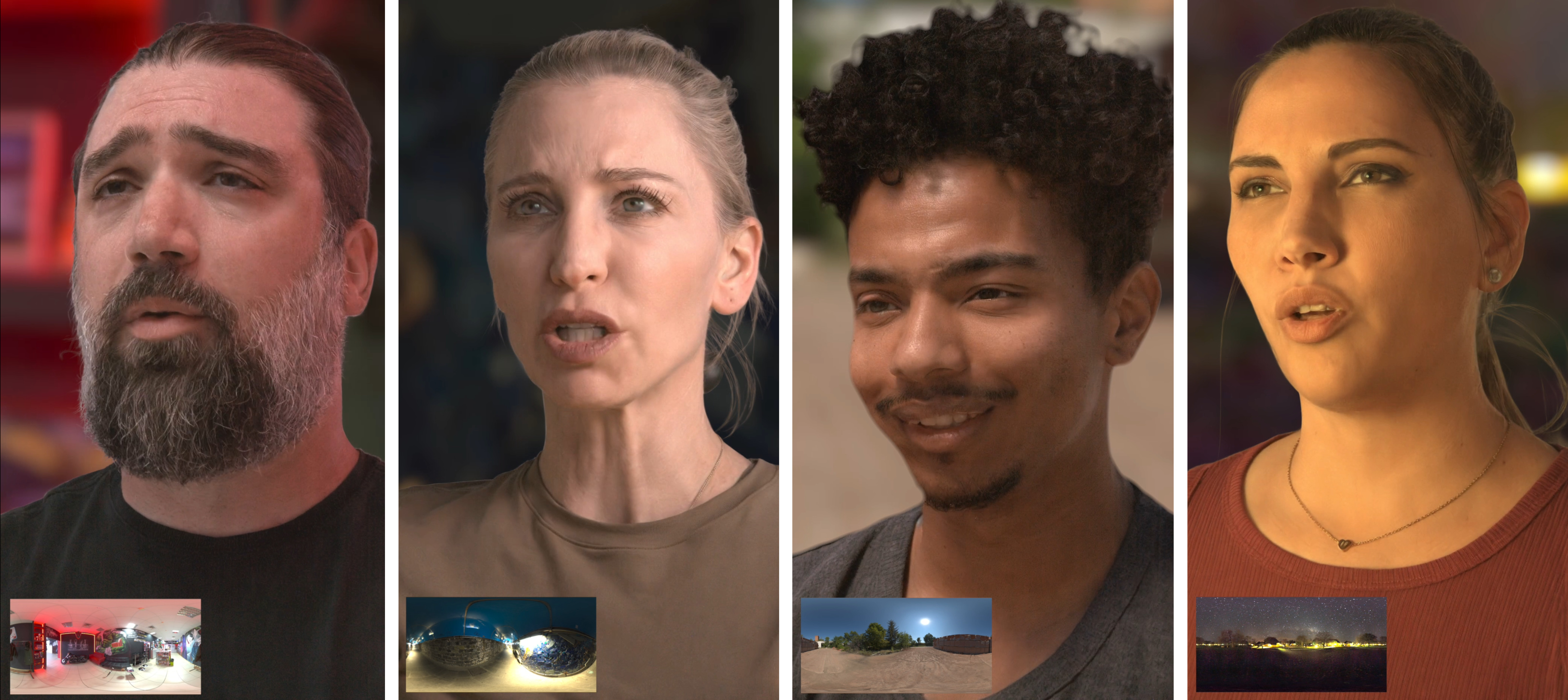}
    \vspace{-2mm}
    \caption{\textbf{HDRI relighting examples}. We can relight a subject under any HDR environment by initially incorporating the HDRI into multiple OLATs or area lights. Then, we relight the subject individually using each OLAT through our diffusion-based relighting method, before compositing the final result in the image space.}
    \vspace{-4mm}
    \label{fig:novellights_hdrs}
\end{figure}

We showcase the effectiveness of our diffusion-based relighting method on the testing data of the four captured subjects. The visual results in Fig.~\ref{fig:teaser}, Fig.~\ref{fig:novelposes}, and Fig.~\ref{fig:arealights} demonstrate that our method reproduces realistic skin texture and reflectance, eye highlights, and fine hair structures while maintaining the subject-specific identity features. The novel lighting animation creates a realistic interplay of shadows and shading across the face. 
Fig.~\ref{fig:arealights} shows our unified lighting control with both directional and area lights, and as an application to our lighting control, Fig.~\ref{fig:novellights_hdrs} shows the results relit by HDRI environment maps. In addition to varying lighting conditions, expression changes are illustrated in Fig.~\ref{fig:novelposes}. 

\begin{table}[t]
\centering
\small
\begin{center}
\caption{Quantitative comparison on different relighting methods.}
\vspace{-2mm}
\label{table:comparison}
\begin{tabular}{ c | c c c c } 
\toprule
                 \textbf{Novel Light}   & PSNR $\uparrow$ & SSIM $\uparrow$ & LPIPS $\downarrow$ & FLIP $\downarrow$ \\ \hline
{Ours}   &  \textbf{30.29}  &  \textbf{.8212}  &  \textbf{.1750}  &  \textbf{.0825}      \\  
{U-Net-based}   &  23.66  &  .7971  &  .2988  &  .1586     \\ 
{ControlNet-based} &  20.30  &  .6708  &  .2281  &  .2304  \\
\midrule
                   \textbf{Novel Expression} & PSNR $\uparrow$ & SSIM $\uparrow$ & LPIPS $\downarrow$ & FLIP $\downarrow$ \\ \hline
{Ours}   &  \textbf{31.55}  &  \textbf{.8281}  &  \textbf{.1585}  &  \textbf{.0689}  \\  
{U-Net-based}   &  23.86  &  .7916  &  .2887  &  .1479  \\ 
{ControlNet-based} &  20.50  &  .6423  &  .2284  &  .2382 \\
\midrule
                  \textbf{Novel View}  & PSNR $\uparrow$ & SSIM $\uparrow$ & LPIPS $\downarrow$ & FLIP $\downarrow$ \\ \hline
{Ours}   &  \textbf{31.77}  &  \textbf{.8254}  &  \textbf{.1601}  &  \textbf{.0689}  \\  
{U-Net-based}   &  24.51  &  .7907  &  .29095  &  .1397 \\ 
{ControlNet-based} & 20.52  &  .6375  &  .2304  &  .2358 \\
\midrule
                  \textbf{Novel LEV}  & PSNR $\uparrow$ & SSIM $\uparrow$ & LPIPS $\downarrow$ & FLIP $\downarrow$\\ \hline
{Ours}   &   \textbf{30.98}  &  \textbf{.8229}  &  \textbf{.1729}  &  \textbf{.0754}  \\  
{U-Net-based}   &  24.55  & .8010   &  .2961  &  .1440  \\ 
{ControlNet-based} &  20.51  &  .6490  &  .2301  &  .2425 \\

\bottomrule
\end{tabular}
\vspace{-6mm}
\end{center}
\end{table}

\subsection{Comparisons}
Since no existing works employ the same training and testing settings as our method, we establish baselines using two different network structures: a diffusion-based model built on ControlNet~\cite{zhang2023adding} and a U-Net-based model inspired by a state-of-the-art image-based relighting technique of~\cite{pandey2021_total_relighting}. We adapt ControlNet to condition on a 3-channel flat-lit image and a 1-channel diffuse shading map, obtained by creating a diffuse shading map using the dot product between the lighting direction and photometric normals. Photometric normals are computed using a standard method~\cite{ma2007rapid} for subjects captured under gradient illumination.
This model is trained under the same settings as ours at $1080\times1920$ resolution using 8 A100 GPUs for 100K iterations with a batch size of 8. The other baseline, a U-Net-based image-to-image translation network, is directly conditioned on the lighting direction, diverging from the use of intrinsic features (normals, albedo, specular) in the referenced architecture to minimize the negative impact caused by inaccurate intrinsic decomposition on a small dataset. Both models are trained using the same data, and we evaluate all methods across four validation configurations: \textit{Novel Light}, \textit{Novel Expression}, \textit{Novel View}, and \textit{Novel Light+Expression+View (LEV)}, introducing complexities in one or three dimensions. We quantify differences between the relit images and ground truth by computing PSNR, SSIM, LPIPS~\cite{zhang2018unreasonable}, and FLIP~\cite{andersson2020flip}, as detailed in Tab.~\ref{table:comparison}. Visual comparisons of these methods with the ground truth are shown in Fig.~\ref{fig:comparison}.

The quantitative results reveal that the ControlNet-based method underperforms in preserving spatial details and color accuracy, reflected by the lowest PSNR, SSIM, and FLIP scores, because it only leverages sparse spatial information. Moreover, it is less sensitive to color shifts, leading to noticeable color differences. The U-Net-based method aligns better spatially with the flat-lit image but tends to produce blurrier results, which improves pixel-level metrics but diminishes photorealism as measured by LPIPS. Our method achieves the best quantitative outcomes, significantly enhancing lighting accuracy, color fidelity, and overall image quality.

\begin{table}[t]
\centering
\begin{center}
\caption{Quantitative results for the ablations. Our method with shading map condition achieves the best result. However, it requires photometric normal as additional input, which increases the cost of capturing the data. }
\vspace{-2mm}
\begin{tabular}{ l|c|c|c|c } 
\toprule
                    & PSNR $\uparrow$ & SSIM $\uparrow$ & LPIPS $\downarrow$ & FLIP $\downarrow$ \\ \hline
Our final model  & 30.04  &  .8141  &  .1719  &  .0859    \\ 
Ours w/ shading map &  \textbf{30.32}  &  \textbf{.8164}  &   \textbf{.1686} &   \textbf{.0836}    \\ 
Ours w/o pyramid noise   &  26.55  &  .7683  &  .1776  &  .1433  \\
Ours w/o pre-trained  &  28.26  &  .8014  &  .1937  & .1041   \\
\bottomrule
\end{tabular}
\vspace{-2mm}
\label{table:ablation}
\end{center}
\end{table}

\begin{figure}
    \centering
    \includegraphics[width=\linewidth]{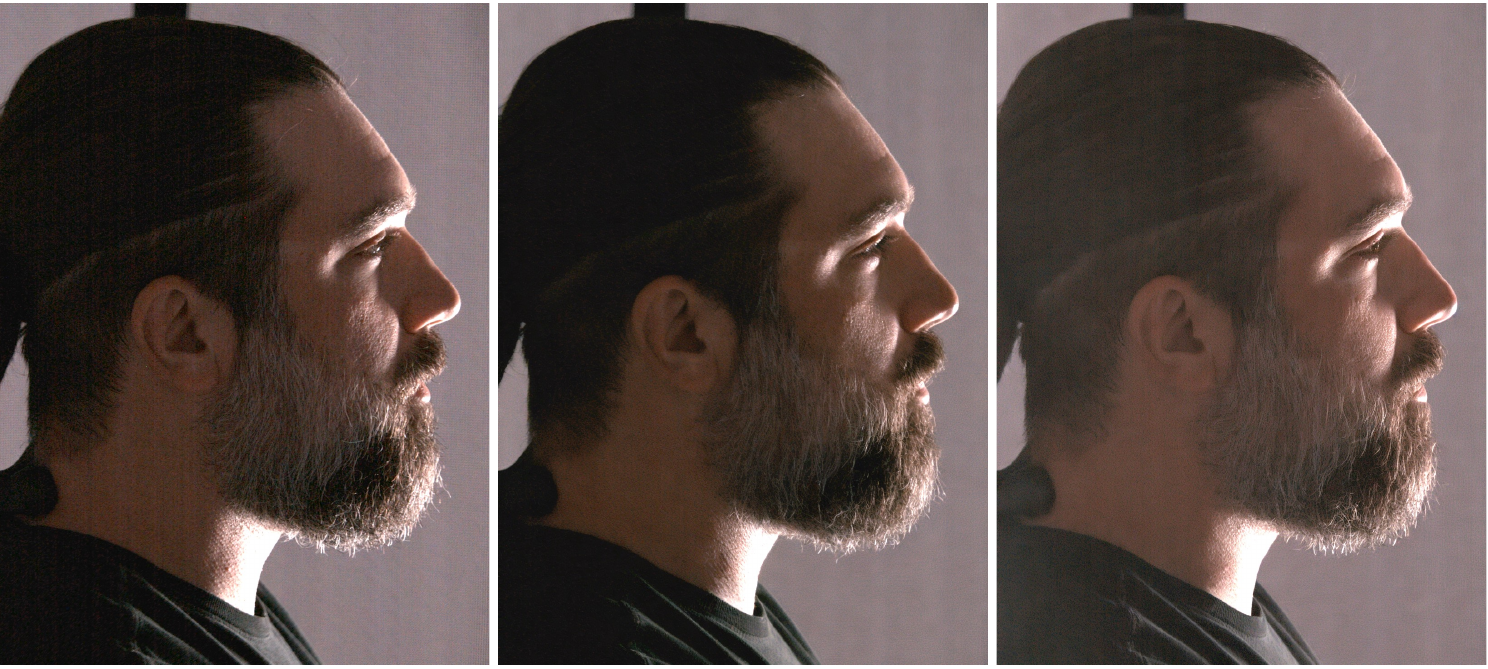}
    \makebox[0.32\linewidth]{\centering \small GT}
    \makebox[0.32\linewidth]{\centering \small Ours}
    \makebox[0.33\linewidth]{\centering \small w/o pyramid noise}
    \vspace{-2mm}
    \caption{\textbf{Visual comparison between using and not using pyramid noise during training.} Using pyramid noise improves the color consistency between the prediction and GT. We increase the brightness of the images by 20\% to exaggerate the error.}
    \label{fig:ablation_pyramidnoise}
    \vspace{-2mm}
\end{figure}

\begin{figure}
    \centering
    \includegraphics[width=\linewidth]{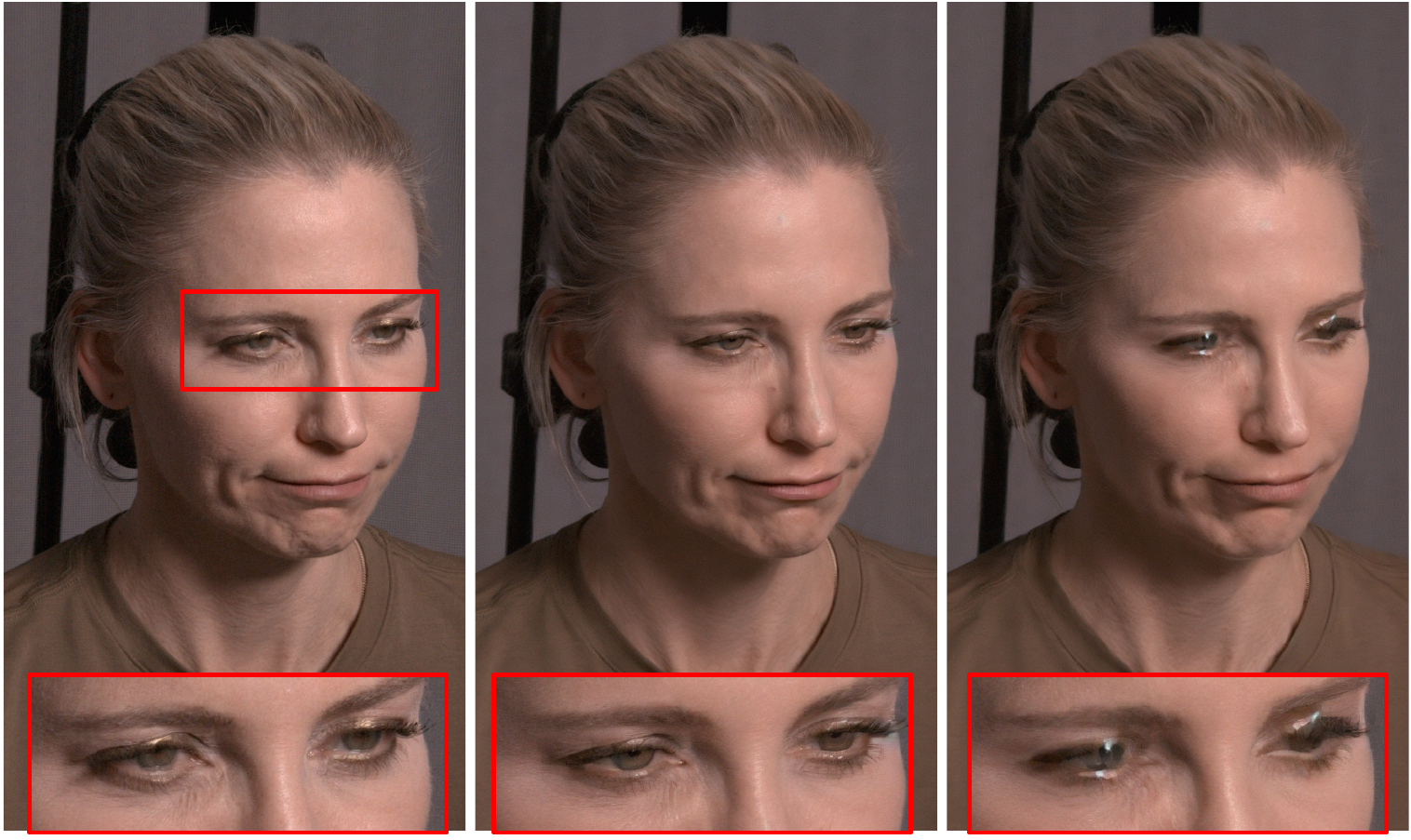}
    \makebox[0.32\linewidth]{\centering \small GT}
    \makebox[0.32\linewidth]{\centering \small Ours}
    \makebox[0.33\linewidth]{\centering \small w/o pre-trained}
    \vspace{-2mm}
    \caption{\textbf{Ablation study on using pre-trained model weights.} We ablate our method by randomly initializing the model weights instead of loading from a pre-trained model (w/o pre-trained). Results show that loading from a pre-trained stable diffusion model helps generate photorealistic results with fewer artifacts.  }
    \label{fig:ablation_pretrained}
    \vspace{-4mm}
\end{figure}

\subsection{Ablation Studies}
\subsubsection{Ablation studies on diffusion-based relighting}
We conducted various ablation studies to evaluate the key technical designs in our diffusion-based relighting model. We first try different methods to integrate lighting control by providing the model with a rough shading estimation. We generate a shading map akin to our strategy in the ControlNet-based relighting model, i.e. dot product between light direction and photometric normal. A visualization of the normal and shading map can be seen in Fig.~\ref{fig:ablation_shading}. We then concatenate the latent of the shading map with the flat-lit latent and random noise as input for the diffusion U-Net. We compare the lighting condition with shading map and SH encoding and results are shown in Fig.~\ref{fig:ablation_shading} and Tab.~\ref{table:ablation}. The numerical results indicate that the shading-map-conditioned model slightly outperforms ours due to the use of its richer shading and geometric detail as spatial control. However, capturing photometric normals in multi-view dynamic settings is challenging due to hardware constraints. Notably, our method does not require time-multiplexed lighting during performance capture, allowing for more comfortable flat lighting and reducing costs by eliminating the need for high-speed global-shutter cameras. Therefore, our approach can achieve comparable performance without relying on photometric normals, making it suitable for more applications.

To determine which factor contributes the most to color fidelity and overall result quality, we disabled the training strategy that uses pyramid noise ("w/o pyramid noise") and removed the initialization that uses the pre-trained Stable Diffusion ("w/o pre-trained"). Quantitative results are detailed in Tab.~\ref{table:ablation}. The model performance noticeably degrades without the pyramid noise, as reflected by a decrease in all evaluation metrics. A visual comparison in Fig.~\ref{fig:ablation_pyramidnoise} further demonstrates that pyramid noise enables more accurate predictions of darker pixels with less color shifting, which likely explains why our method better preserves color fidelity. Additionally, fine-tuning the network based on pre-trained stable diffusion is crucial as it provides strong prior knowledge to generating photorealistic relit results. Without pre-trained weights, the model tends to generate more artifacts, as illustrated in Fig.~\ref{fig:ablation_pretrained}.

\begin{table}[t]
\centering
\caption{Quantitative comparison on different models of dynamic 3DGS.}
\small
\vspace{-2mm}
\begin{center}
\begin{tabular}{ l|c|c|c } 
\toprule
  & PSNR $\uparrow$ & SSIM $\uparrow$ & LPIPS $\downarrow$ \\ \hline
Baseline model~\cite{jung2023_deformable_gs} & 30.202 & 0.876 & 0.254 \\ 
(a) Partition w/o K-frame & 31.717 & 0.893 & 0.239 \\ 
(b) Partition w/ K-frame & 31.114 & 0.901 & 0.218 \\ 
(c) Partition w/ K-frame \& $L_{reg}$ (Ours) & \textbf{31.979} & \textbf{0.903} & \textbf{0.216} \\ 
\bottomrule
\end{tabular}
\end{center}
\label{table:ibr_ablations}
\vspace{-2mm}
\end{table}

\subsubsection{Ablation studies on scalable dynamic 3D Gaussian Splatting}
To demonstrate the effectiveness of our scalable dynamic 3DGS, we compare it with three baselines: (1) the original deformable 3DGS model~\cite{jung2023_deformable_gs}; (2) a model partitioning long sequences for separate training without our K-frame initialization (K-frame) or deformation offset regularization ($L_{reg}$); and (3) a model using partitioning and K-frame but without $L_{reg}$. We evaluate by holding out 10\% of the captured views from one subject’s dynamic sequences, using quantitative metrics. Tab.~\ref{table:ibr_ablations} shows improvements in reconstruction quality and consistency with the proposed strategies.

\begin{figure}
    \raggedleft
    \includegraphics[width=1.0\linewidth]{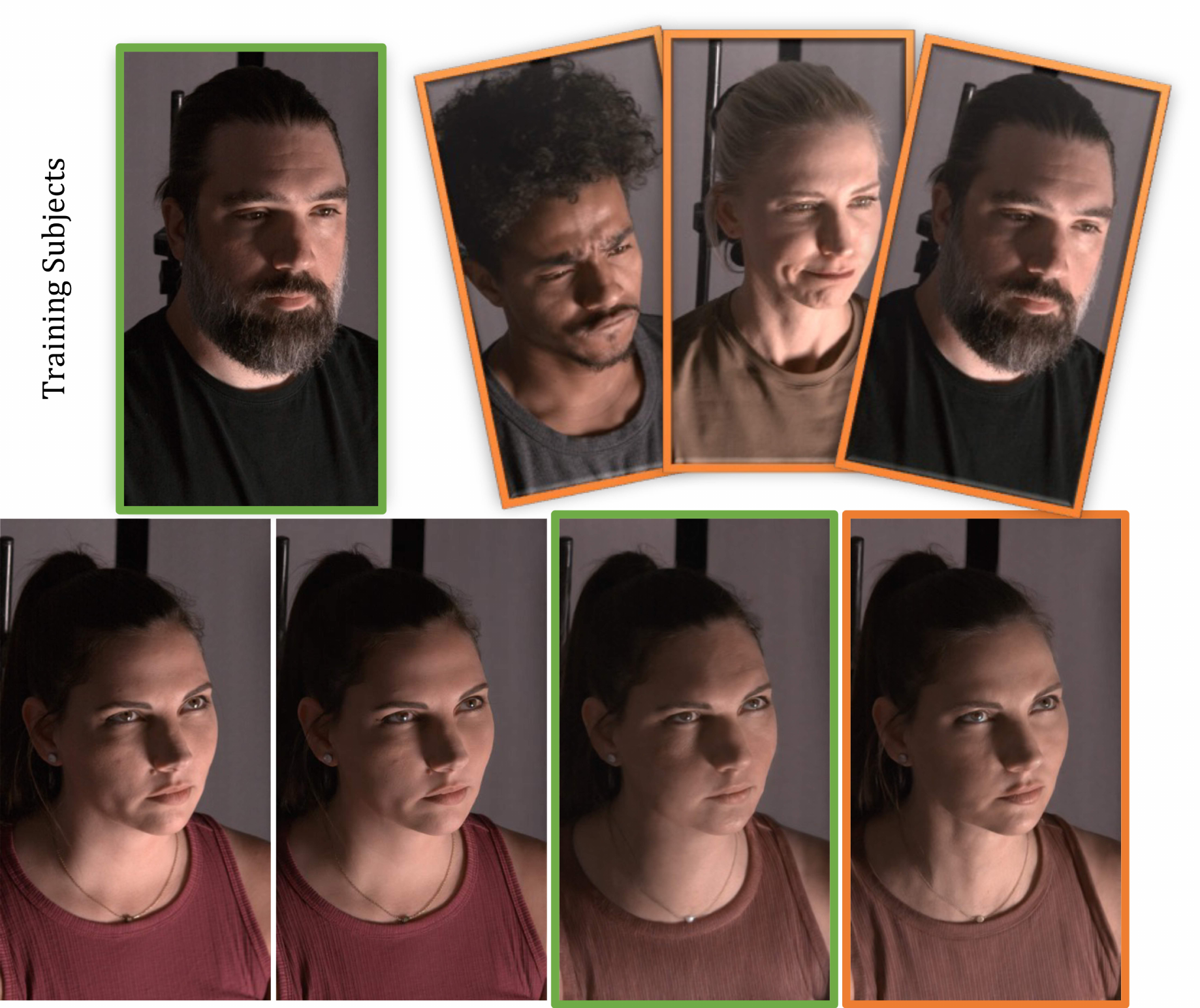}
    \vspace{-2mm}
    \makebox[0.23\linewidth]{\centering \small Ground Truth}
    \makebox[0.23\linewidth]{\centering \small Subject-specific}
    \makebox[0.23\linewidth]{\centering \small Single-subject}
    \makebox[0.23\linewidth]{\centering \small Three-subject  }
    \makebox[0.045\linewidth]{\centering \small  }
    \vspace{-2mm}
    \caption{\textbf{Generalization capacity.} We test the generalizability of our model by evaluating on an unseen subject. Our subject-specific model best preserves identity when trained and tested on the same subject. Interestingly, the single-subject model can relight a new subject but may inherit features like skin texture and lip color from the training subject. If trained on three other subjects, the model can better preserve the new subject's unique features but still shows noticeable changes in color and fine details.}
    \label{fig:ablation_crossid}
    \vspace{-6mm}
\end{figure}

\section{Real-World HDRI Relighting}
Since our method infers OLAT images from any viewpoint that closely match the ground truth, it can reproduce real-world lighting conditions from HDRI maps, following~\cite{debevec2000_acquiring_reflectance_field}. To evaluate this, we capture reference images of one trained subject in complex indoor and outdoor environments, using HDRI maps to reconstruct the lighting. For comparison, we render the flat-lit subject from similar viewpoints and input these into our relighting model. For each HDRI map, we generate two results: one using the OLAT-based model and the other using the area-light model. In OLAT-based relighting, we linearly combine the results of 123 OLATs, while in area-light relighting, we approximate the HDRI map using 15 Spherical Gaussians (SGs), infer for each SG, and then combine their results. To account for spectral differences between LED and real-world lighting, we capture a color chart in both environments and compute a constant 3-channel scaling factor to correct the relit image. Fig.~\ref{fig:real_world_hdri} shows that our results are comparable to reference images under complex lighting conditions.

\section{Model Generalization}
Although our model is primarily trained on the subject-specific data, which limits its generalization capacity for novel subjects, the model learns to translate from flat to directional lighting, a process that is somewhat subject-agnostic. However, subject-specific information from the model, especially high-frequency details, may remain in the results, as in Fig.~\ref{fig:ablation_crossid}. To assess whether training on multiple subjects' data reduces such subject-specific information, we train a model using data from three subjects. This model shows improved performance on a novel subject, better preserving the subject's identity, though color shifts are still prominent. This suggests the potential for training a general relighting model using more subjects.

%% file: limitation.tex
\begin{figure}
    \centering
    \includegraphics[width=\linewidth]{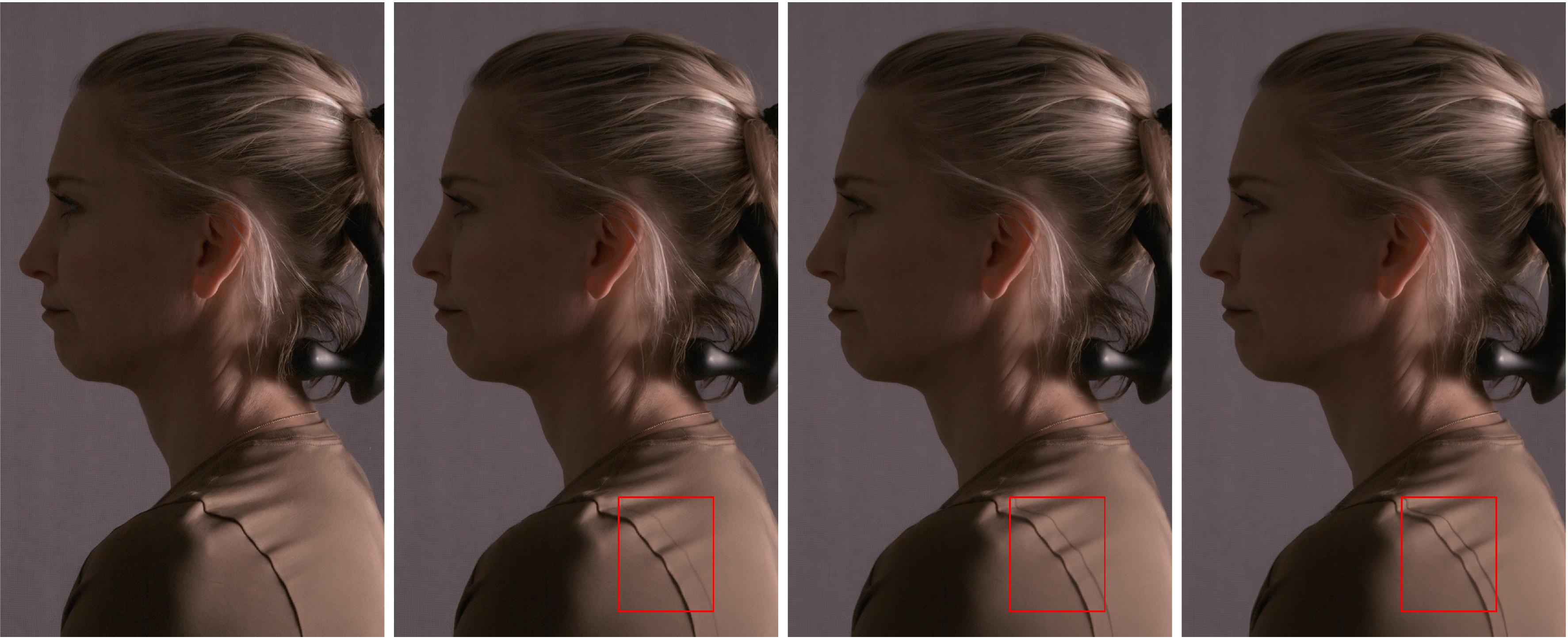}
    \makebox[0.242\linewidth]{\small Raw Output}
    \makebox[0.242\linewidth]{\small Frame 1}
    \makebox[0.242\linewidth]{\small Frame 2}
    \makebox[0.242\linewidth]{\small Frame 3}
    \vspace{-5mm}
    \caption{\textbf{Limitation.} The post-processing keyframe interpolation may cause artifacts, such as double strips on shirts, due to inaccurate warping from fast camera motion, which affects the precision of optical flow estimation.}
    \label{fig:enter-label}
    \vspace{-5mm}
\end{figure}

\section{Limitations} 
Our technique has a few limitations; for one, we do not completely resolve the temporal consistency issue of image-based diffusion models due to the absence of video training data, and our optical flow post-processing occasionally introduces artifacts during rapid movements, as illustrated in Fig.~\ref{fig:enter-label}. Emerging video diffusion models hold promise for solving this issue. Secondly, our subject-specific training is not designed to generalize to unseen subjects and can alter identity features in the relit results, as shown in our extension. We believe this limitation could be addressed by training on a diverse, multi-person dataset and/or implementing identity disentanglement techniques. Lastly, we would need to add full-body training data for our relighting technique to be applied to full-body performances.

%% file: conclusion.tex
\section{Conclusion}

The proposed method shows promising results in relighting free-viewpoint flat-lit facial performances to various lighting conditions using subject-specific OLAT training data. It demonstrates enhanced performance in replicating eye highlights, skin luster, and hair details compared to prior image-based methods, which start with a single diffuse lighting condition. Leveraging an underlying diffusion model allows us to achieve well-conditioned results while maintaining the original subject's appearance and identity by finetuning on subject-specific data. These findings suggest a promising direction for advancing relightable volumetric capture production and facilitating postproduction relighting of any flat-lit footage.

%% file: full_results_figures_2pages.tex
\begin{figure*}
    \centering
    \includegraphics[width=0.95\linewidth]{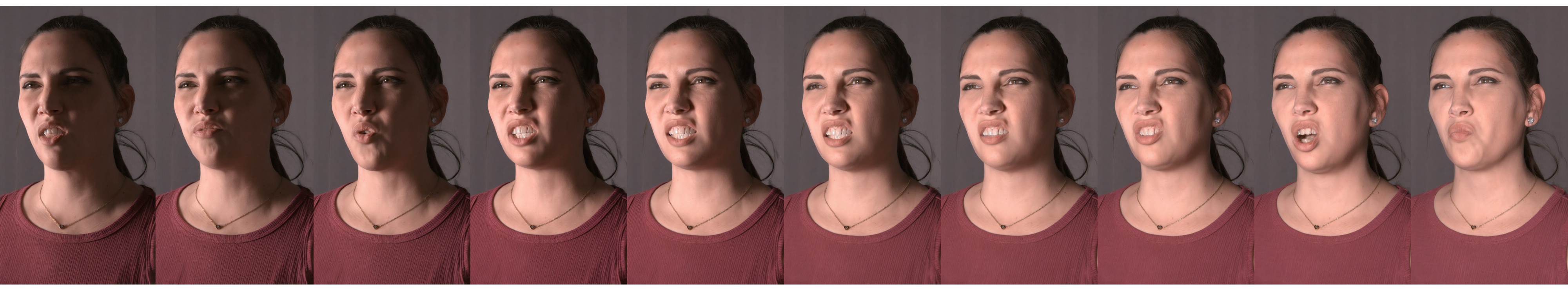}
     \vspace{-3mm}
    \caption{\textbf{Novel facial expressions under a rotated directional lighting.} Each row shows every 12th frame of a performance sequence of 120 frames for an individual actor, having a novel facial expression every frame. }
    \label{fig:novelposes}
\end{figure*}

\begin{figure*}
    \centering
    \includegraphics[width=0.95\linewidth]{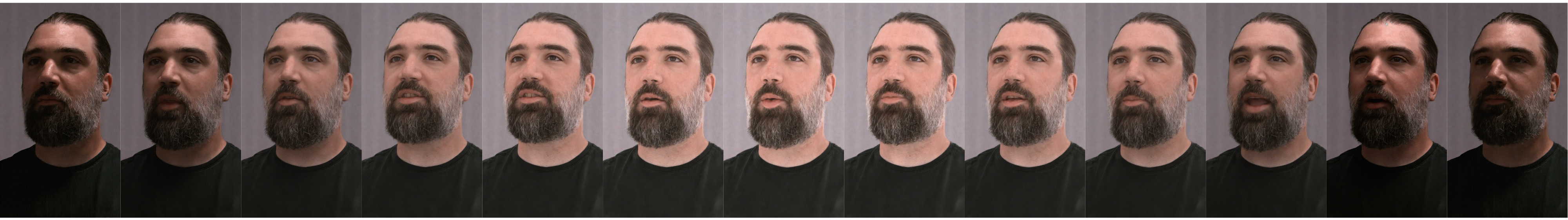}
     \vspace{-3mm}
    \caption{\textbf{Unified lighting control combining directional and area lights.} The image sequence shows an animated area light on a performance sequence with a fixed view. The area light animates linearly back and forth between a directional light (0 degree area light) and a 360 degree area light. In order to compute the reflectance fields of an area light, weighted captured OLAT images are linearly combined. }
    \label{fig:arealights}
\end{figure*}

\begin{figure*}
    \centering
    \includegraphics[width=0.95\linewidth]{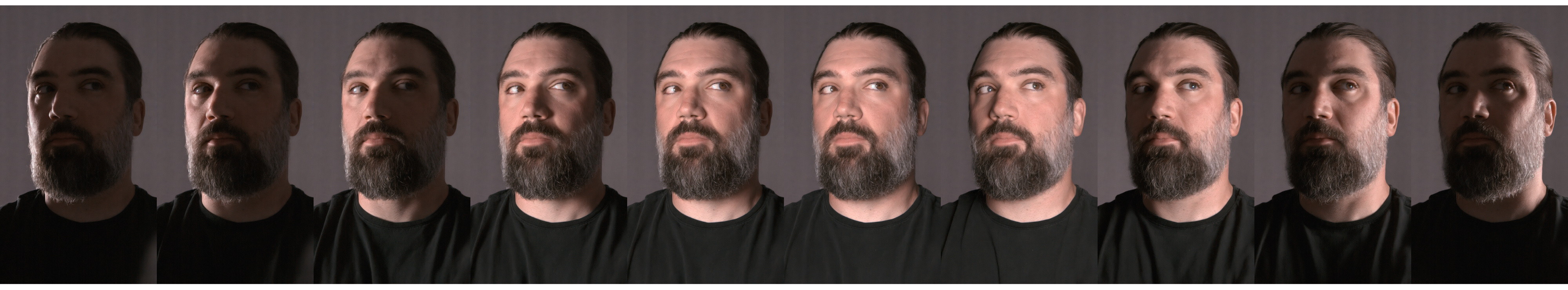}
    \vspace{-5mm}
    \caption{Subject seen under novel directional lightings, following a horizontal 180 degree rotational path.}
    \label{fig:novellights}
\end{figure*}

\begin{figure*}
    \centering
    \includegraphics[width=0.95\linewidth]{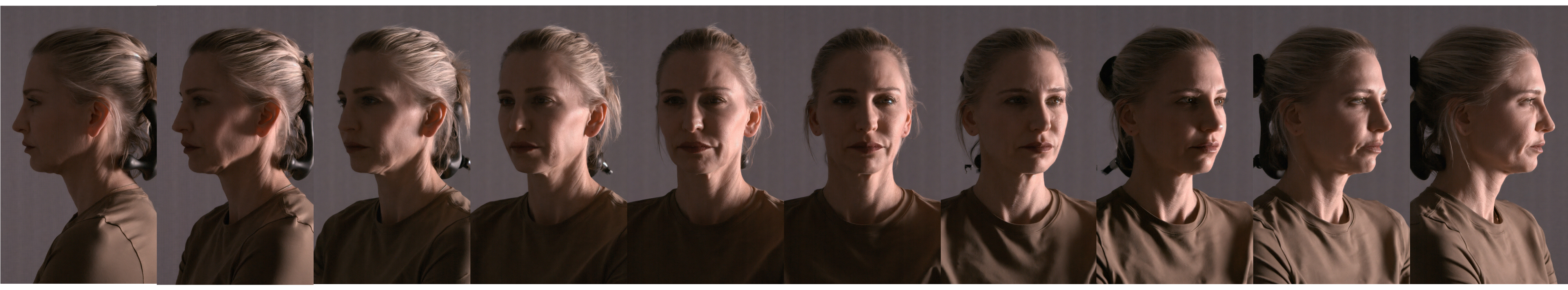}
    \vspace{-5mm}
    \caption{Subject seen under 10 novel views, following a 180 rotation around the subject. }
    \label{fig:novellights}
\end{figure*}

\begin{figure*}
    \centering
    \includegraphics[width=0.95\linewidth]{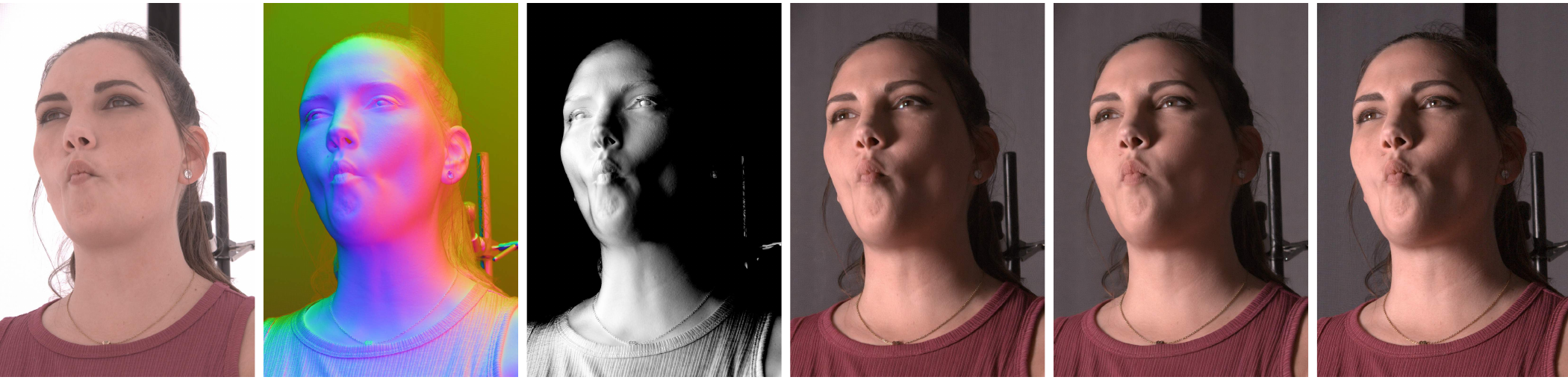}
    \makebox[0.16\linewidth]{\centering \small Input}
    \makebox[0.16\linewidth]{\centering \small Photometric normal}
    \makebox[0.16\linewidth]{\centering \small Shading map}
    \makebox[0.16\linewidth]{\centering \small Ours w/ shading map}
    \makebox[0.16\linewidth]{\centering \small Ours w/ SH encoding}
    \makebox[0.16\linewidth]{\centering \small GT}
    \caption{\textbf{Comparisons between different lighting condition methods, shading map and SH encoding.} We also visualize the photometric normal and shading map, where the shading map is the dot product between photometric normal and light direction. The two methods achieve comparable visual quality, whereas the method conditioned on the shading map requires the photometric normal as additional input. }
    \label{fig:ablation_shading}
\end{figure*}

\begin{figure*}[t]
    \centering
    \includegraphics[width=\textwidth]{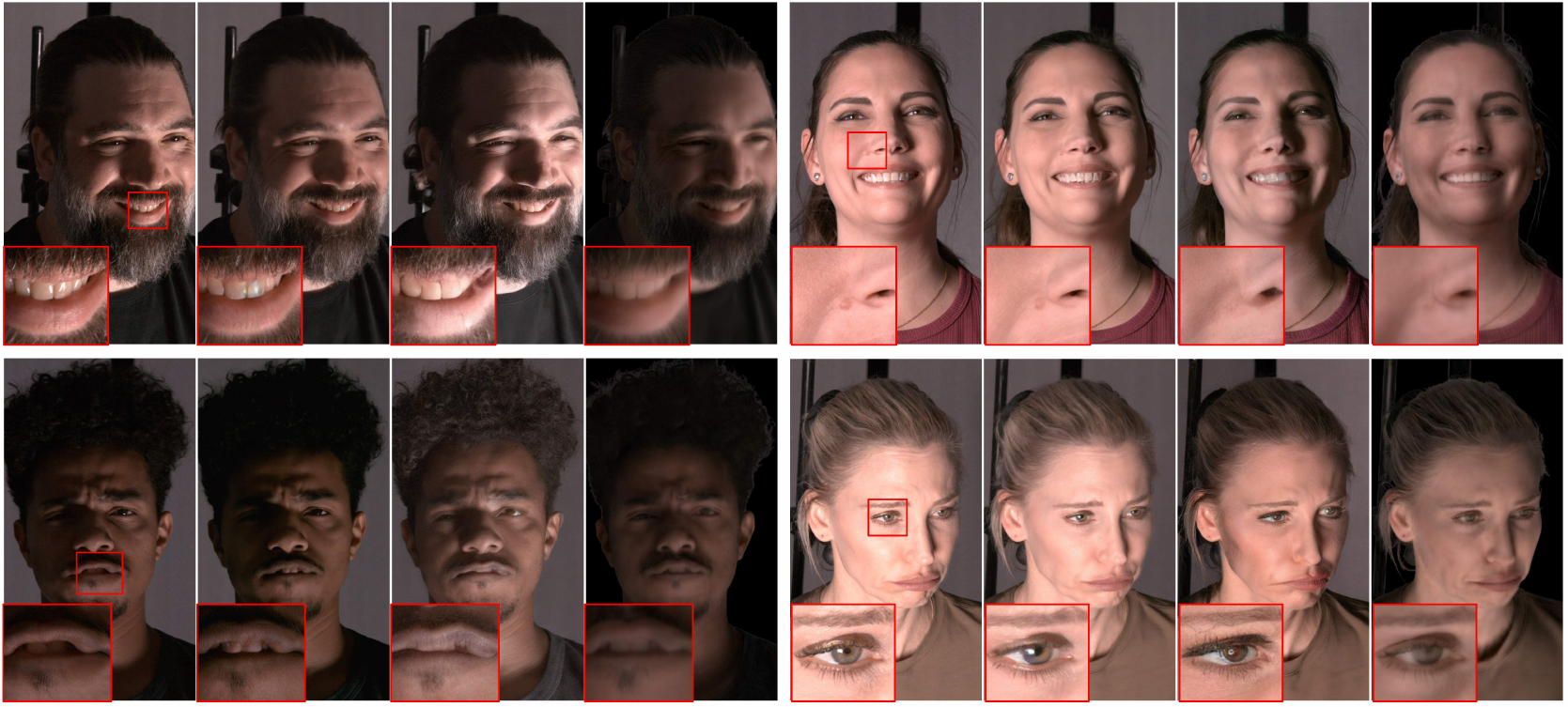}
    \makebox[0.12\linewidth]{\centering \small GT}
    \makebox[0.12\linewidth]{\centering \small Ours}
    \makebox[0.12\linewidth]{\centering \small ControlNet-based}
    \makebox[0.12\linewidth]{\centering \small U-Net-based}
    \makebox[0.12\linewidth]{\centering \small GT}
    \makebox[0.12\linewidth]{\centering \small Ours}
    \makebox[0.12\linewidth]{\centering \small ControlNet-based}
    \makebox[0.13\linewidth]{\centering \small U-Net-based}
    \vspace{-5mm}
    \caption{\textbf{Comparison of the results generated by ControlNet-based relighting, U-Net-based relighting, and our method}. Note that the U-Net-based model is trained with foreground masks to focus on foreground relighting and generate black background while the other two methods never use masks for training; they are instead generated using 3DGS. We remove the backgrounds from the results of other methods when calculating metrics.}
    \label{fig:comparison}
\end{figure*}

\begin{figure*}
    \centering
    \includegraphics[width=\linewidth]{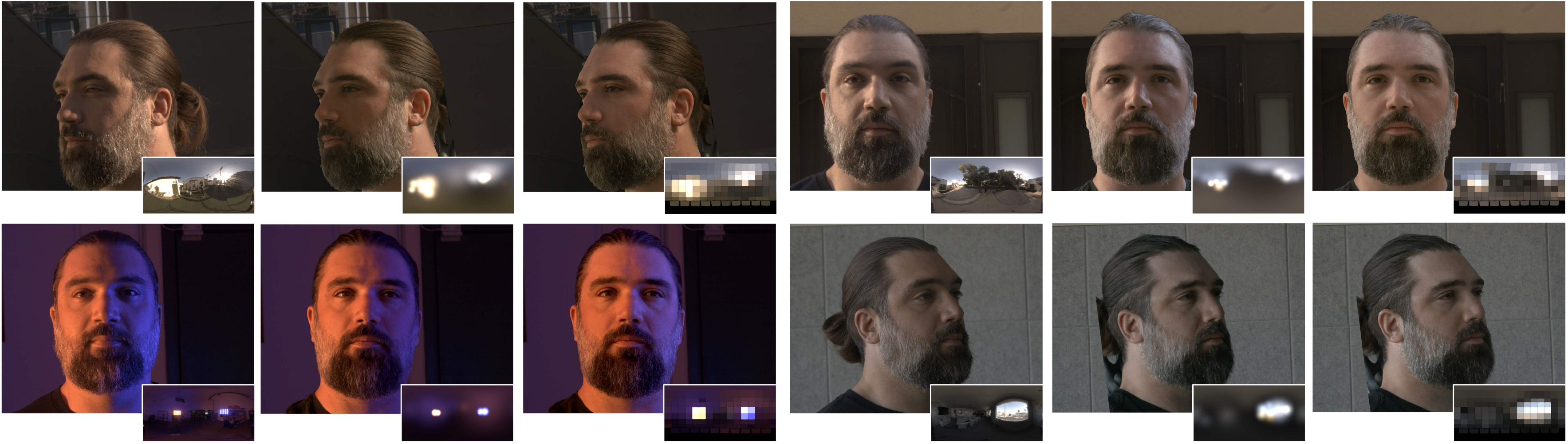}
    \makebox[0.16\linewidth]{\centering \small Reference}
    \makebox[0.16\linewidth]{\centering \small Area-light model}
    \makebox[0.16\linewidth]{\centering \small OLAT-based model}
    \makebox[0.16\linewidth]{\centering \small Reference}
    \makebox[0.16\linewidth]{\centering \small Area-light model}
    \makebox[0.16\linewidth]{\centering \small OLAT-based model}
    \caption{\textbf{Real-world HDRI relighting}. We compare the HDRI relighting results from our method with reference images captured in four real-world environments. For each HDRI relighting example, we present the reference image, two relighting results (one using the area-light model and the other using the OLAT-based model), along with the captured HDRI map and its approximations of 15 Spherical Gaussians (SGs) and 123 OLATs overlaid on the image. Both models achieve results comparable to the reference images.}
    \label{fig:real_world_hdri}
\end{figure*}

%% file: appendix.tex
\section{Appendix}

\subsection{Method}
\subsubsection{3D Gaussian Splatting Reconstruction with Background Segmentation}

As some post-composition like the generation of HDR environment lighting or training data of our U-Net based network necessitates foreground matting, we propose employing the clean background captured by our stage as the known background in the 3DGS reconstruction. Implementation details are provided here.

In the 3D Gaussian Splatting reconstruction, we start with a set of multi-view images, along with corresponding camera calibration and sparse point cloud generated by Metashape ~\footnote{https://www.agisoft.com/}. Following~\cite{kerbl2023_3d_gs}, we initialize the 3D Gaussians with the sparse point cloud and then optimize the Gaussians, which are defined by a center position $\textbf{x}$, a covariance matrix $\Sigma$ obtained from rotation $\textbf{r}$ and scaling $\textbf{s}$, opacity $\alpha$, and color $\textbf{c}$ represented by spherical harmonic coefficients. Given a view direction transformation matrix $\textit{W}$, the convariance matrix $\Sigma$ can be projected to camera space following~\cite{zwicker2002ewa} and converted to:
\begin{equation}
    \Sigma' = J W \Sigma W^T J^T,
\end{equation}
where $J$ is the Jacobian of the affine approximation of the projective transformation and $\Sigma$ is expressed as follows:
\begin{equation}
    \Sigma = RSS^TR^T
\end{equation}
with the rotation and scaling vectors $\textbf{r}$ and $\textbf{s}$ transformed into matrices $R$ and $S$.
The color of each pixel $p$ on the image is defined as alpha blending of 3D Gaussians that covers this pixel and are sorted in depth:
\begin{equation}
\label{eq:gs_color}
    \begin{split}
    c_{p} & = \sum_{i \in N}T_i \alpha_i c_i \\
    \alpha_i & = \sigma e^{\frac{1}{2} (p-q_i)^T \Sigma' (p-q_i)},
    \end{split}
\end{equation}
where $T_i$ is the transmittance defined as $T_i = \prod^{i - 1}_{j = 1} (1 - \alpha_j)$ and $q_i$ is the 2D projection position of the $i$th Gaussian. To optimize Gaussians, the loss function is L1 combined with a D-SSIM term same as the original loss in~\cite{kerbl2023_3d_gs}.

\begin{figure}[t]
    \centering
    \includegraphics[width=1.\linewidth]{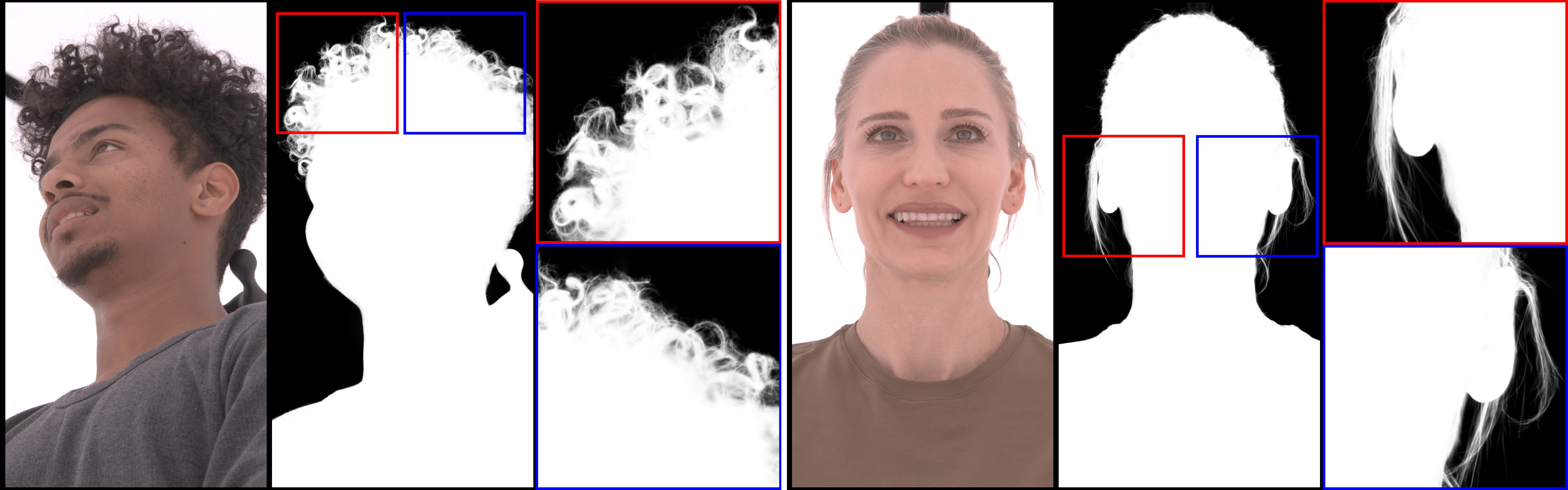}
    \caption{Alpha mattes produced by our 3D GS reconstruction utilizing the clean plate. Each example includes the input image along with the generated alpha mattes. Zoomed-in details highlight specific elements, such as hair strands, that are distinctly segmented through the use of the clean plate.}
    \label{fig:method_mask}
\end{figure}

To obtain precise alpha mattes for the foreground subject, we employ the captured clean plate as a known background. This facilitates the separation of foreground and background Gaussians, directing the optimization to focus only on the foreground Gaussians. Therefore, the final pixel color is computed as:
\begin{equation}
    c'_{p} = \sum_{i \in N}T_i \alpha_i c_i + (1 - \sum_{i \in N}T_i \alpha_i) b_p,
\end{equation}
where $b_p$ is the pixel color on the clean plate. Fig.~\ref{fig:method_mask} shows the examples of high-quality alpha mattes of the reconstructed foreground subject by 3D GS, aided by the clean plate.

\subsection{Additional Results}
\begin{figure}[t]
     \centering
     \includegraphics[width=1\linewidth]{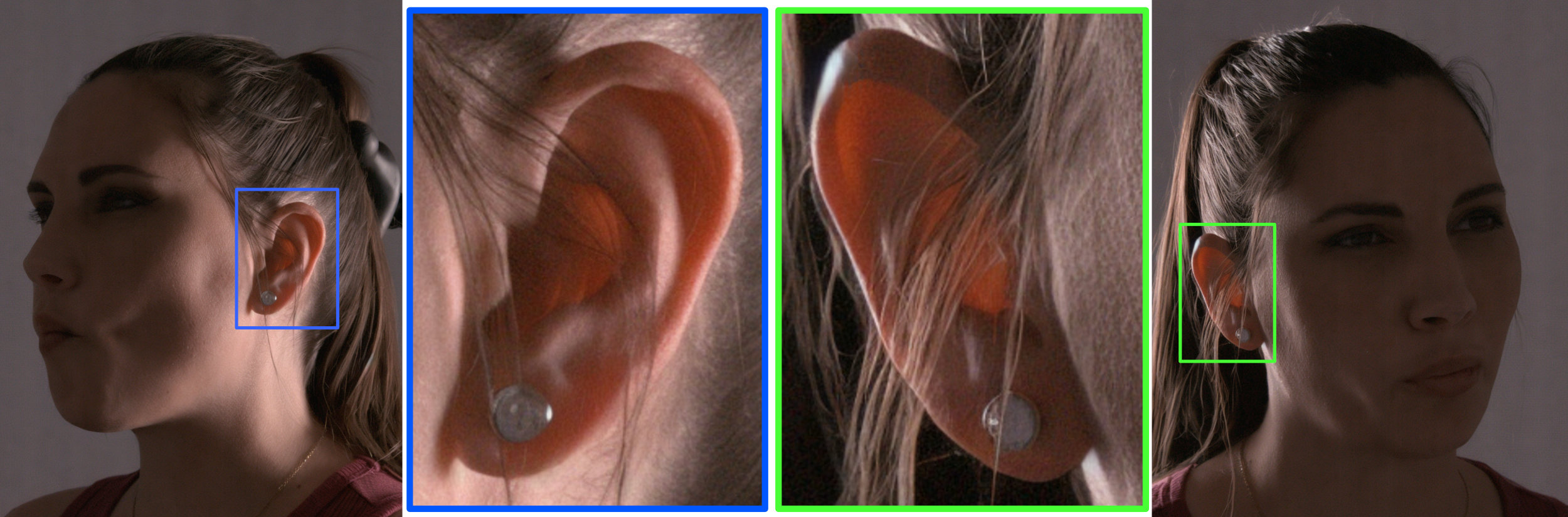}
     \caption{Complex light transport effects, showing scattering and transmitting light through translucent ear together with sharp specular reflection at grazing angles and subtle reflectance of hair strands, achieved with our method on unseen views and light directions. }
     \label{fig:complex_shading_effects}
\end{figure}

Fig.~\ref{fig:complex_shading_effects} shows complex light transport effects of our DifFRelight method. The complicated lighting effects partially originate from the stable diffusion prior, but most are learned by our model during the fine-tuning process from subject specific real-world reflectance data applying various lighting conditions. Accurately modeling these effects with physically based rendering is extremely challenging and requires complex computational models to simulate light transmittance and subsurface scattering in multi-layered materials. For example, the human skin itself consists of multiple layers—epidermis, dermis, elastic cartilage, and fatty tissue—that scatter light diffusely. Additionally, various sized blood vessels pervade the ear tissue, which absorbs light in the red spectrum producing the reddish hue from transmitting light.

\subsection{Comparison}
Next we provide the implementation details of the U-Net based relighting baseline and add two new comparisons with zero-shot relighting methods.
\subsection{U-Net Training Pipeline}
To train our U-Net~\cite{Ronneberger2015_unet} based architecture for a comparison, similar to the diffusion based training pipeline, we use pairs of flat-lit captured images and captured OLAT images, along with camera space light direction as input matching the capture. In addition we also made use of the masks we acquire from the GS reconstruction to help our model skip learning the background. This helps convergence as the transform between the flatlit image and the target background is more ambigious than the foreground subject. Encoder architecture of the U-Net is based on the ResNet-50~\cite{he2015deepres} architecture. Matching this architecture allows weight initializations from a pre-trained model, however we elected to not use these weights as frozen. We observed that convergence is relatively faster with this initialization strategy, but a random weight initialization yields a similar convergence at longer training times. For the decoder, we use style modulations described in StyleGAN2~\cite{karras2020styleGANQuality}. Style code is used for light directions, where a multi-layer perceptron with two hidden layers with 128 neurons each hidden layer, map three dimensional light directions into the shape necessary per CNN layer of the decoder, for modulations described in the StyleGAN2 paper. Decoder weights are not initialized from a StyleGAN2 checkpoint but instead initialized randomly. We train this model per performer for about 30 epochs for convergence and it takes considerably more time to converge than the diffusion based training method. Multi GPU training is used on 4 NVIDIA A10G cards, with 2k crops from the original 4k resolution. Twenty percent of the dataset is held matching our other training configurations for comparison. Training takes multiple weeks compared to less than a day, for considerably inferior outputs. Different to the diffusion based training pipeline, we use the mask from the GS reconstruction to take out the background before the L2 norm between target OLAT image from the capture and the prediction. Adam optimizer is used for optimization with a learning rate of 3e-4. For reconstructing the eye highlights increasing the exposure by multiplying the intensity of the target image by two helped getting better results. Inference times are interactive and it was possible to build an interactive application that enables the user to pick light directions as the model shows predictions at interactive speeds of 16 to 24 frames per second on playback with commercial CUDA enabled graphics cards. However this improvement in inference time isn't justified considering the quality of results and the training times needed for convergence. We believe better results may be obtained given more variation in data and with a bigger dataset to build a prior with this method. However Stable Diffusion based method already has a prior we could employ and the diffusion based model seems to be more expressive for high frequency details. U-Net method also suffers from temporal stability where the model is not as confident about the shadow shapes or placement of some highlights. We believe this may improve with denser capture configurations of lights and cameras on our stage, which would also make the capture process more expensive. Further exploration using adversarial loss is left out, as this training method is harder to tune for reasonable convergence during training and it would not make sense for a production tool where we may need to make this work per performer in a reproducible manner without much supervision.

\begin{figure}[t]
    \centering
    \includegraphics[width=\linewidth]{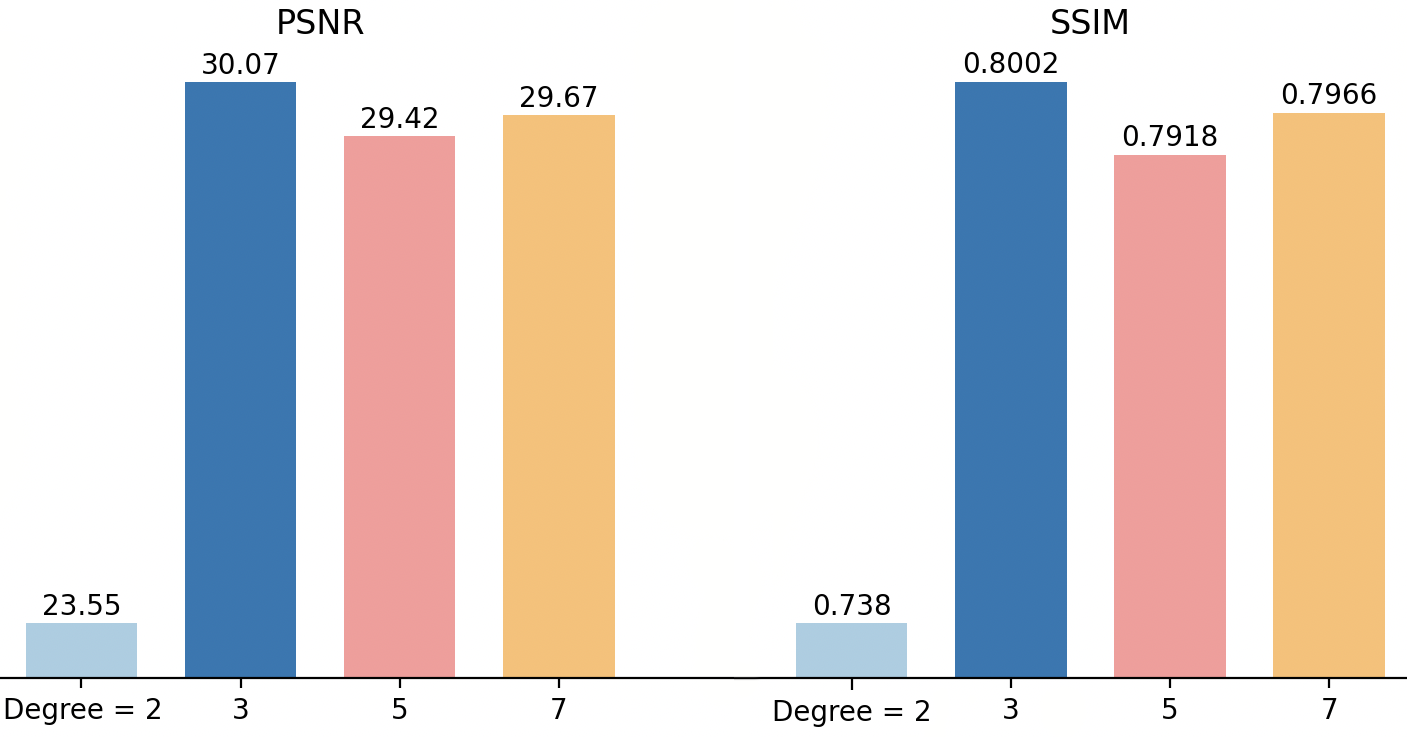}
    \caption{We test the affect of different SH degrees to the PSNR and SSIM of the relit results. We use degree $= 3$ in all of other experiments. }
    \label{fig:ablation_sh}
\end{figure}

\subsection{Comparison with Zero-Shot Relighting Methods}

\begin{figure*}[t]
    \centering
    \includegraphics[width=0.95\linewidth]{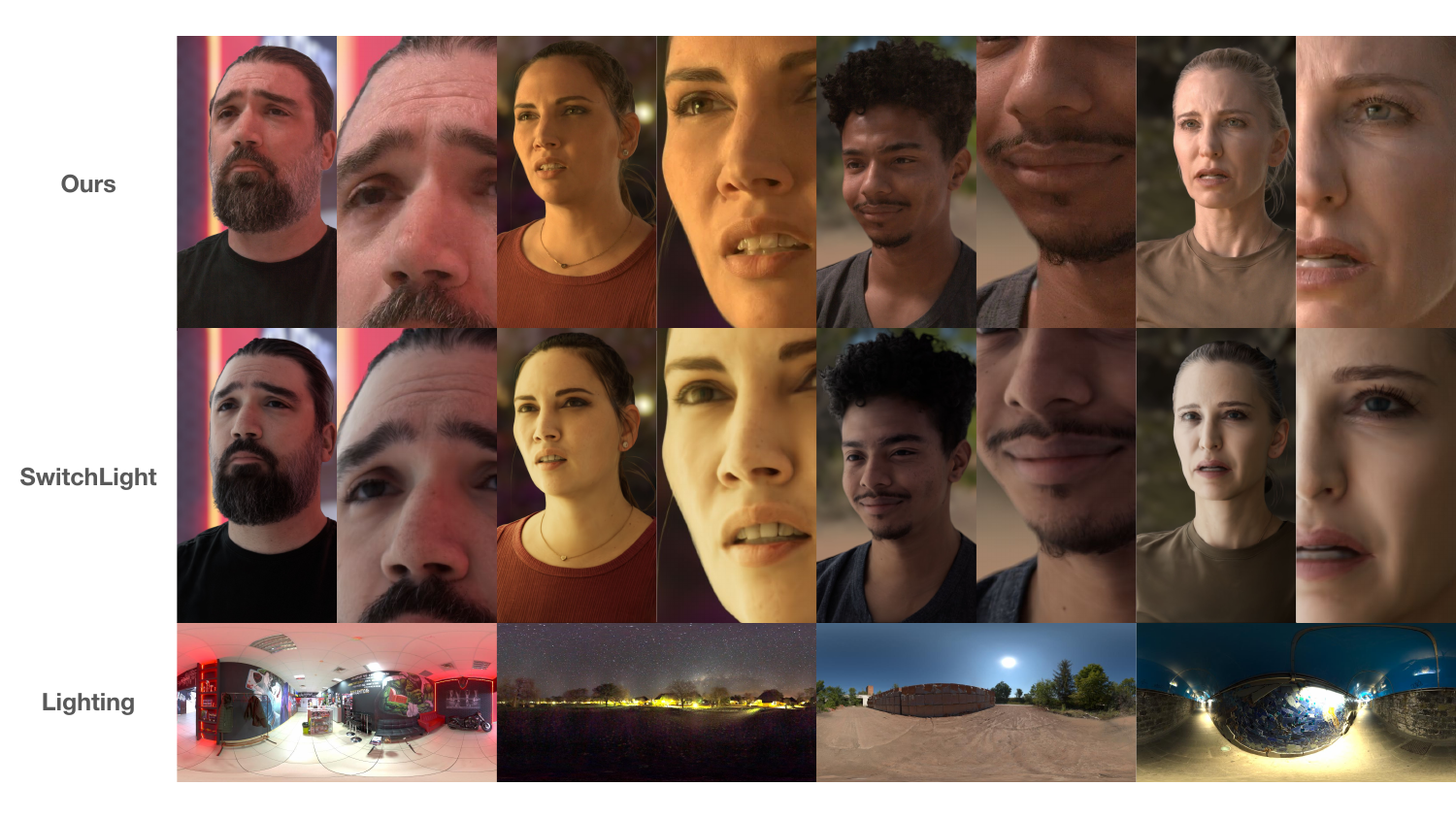}
    \caption{A comparison between our method and the method \citet{kim2024switchlightcodesignphysicsdrivenarchitecture}, implemented in the product named SwitchLight Studio, with the last row showing the HDRI environment lighting.
    }
    \label{fig:switchlight}
\end{figure*}

We first provide a comparison with SwitchLight~\cite{kim2024switchlightcodesignphysicsdrivenarchitecture}, one of the most recent state-of-the-art traditional relighting methods, in Fig. \ref{fig:switchlight}, showing results from our method along with results generated using the same lighting information and input flat-lit image. Due to finetuning with subject specific real data and not relying on a rendering model to produce final imagery, generated skin colors with our method look more natural and detailed, as well as specular reflections on skin and eyes are a bit more distinctive. The physically based rendering approach in Switchlight emphasizes the difficulty in reproducing the color tone of the HDRI skin reflections in a natural way. Learning a general model to reconstruct geometric properties as done in Switchlight is a great challenge and comes with caveats in normal and albedo quality. Though, SwitchLight in comparison generalizes to any subject and can be used as a general relighting tool.

Recently, IC-Light \cite{iclight} was released as a zero-shot relighting algorithm based on Stable Diffusion. It can harmonize the lighting of a foreground image with a given background, where the background controls the lighting of the output image. 

\begin{figure*}[t]
    \centering
    \includegraphics[width=0.95\linewidth]{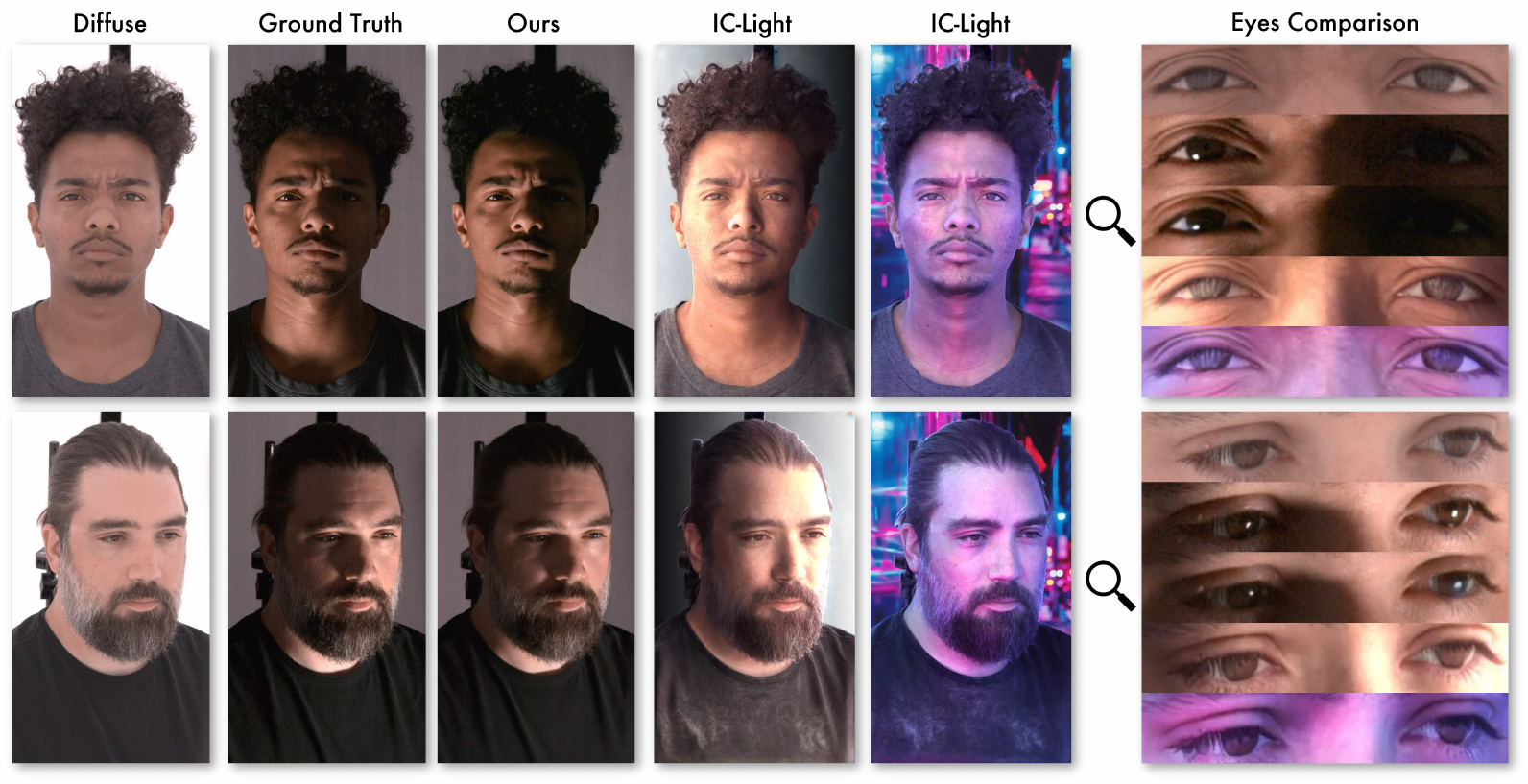}
    \caption{A comparison between our method and IC-Light \cite{iclight}. IC-Light fails to produce highlights and reflections on eyes, and forehead wrinkles are completely absent. The official implementation's default setting for relighting from the left was used to produce the first IC-Light image, and the included cyberpunk background and prompt were used for the rightmost image. }
    \label{fig:iclight}
\end{figure*}

\begin{figure*}
    \centering
    \includegraphics[width=0.95\linewidth]{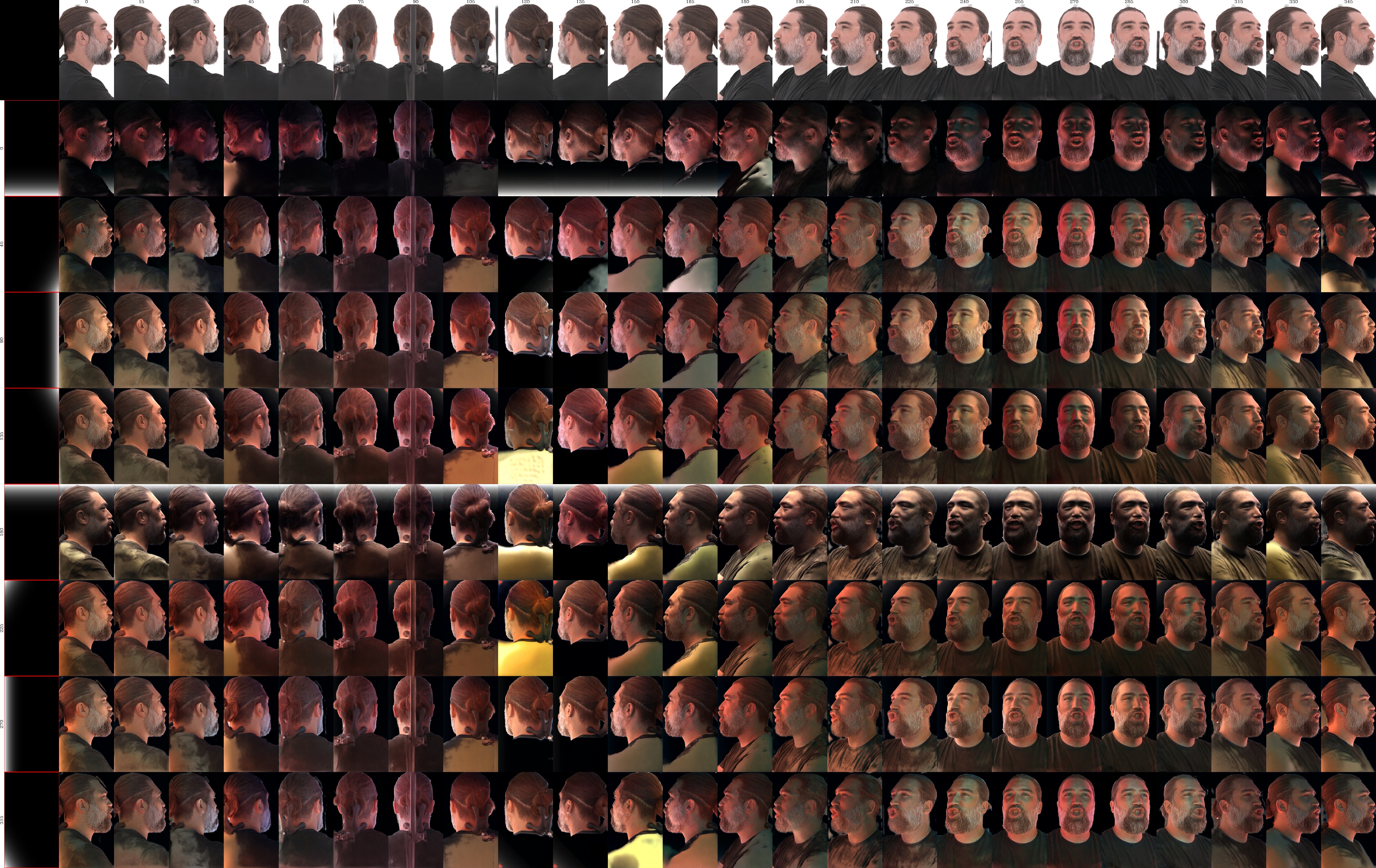}
    \caption{IC-Light is temporally inconsistent, and not able to light the subject precisely. The y-axis is the angle of the light, and the leftmost column is the background image used. The x-axis represents the angle in which our subject is rotated in 3d.}
    \label{fig:icgrid}
\end{figure*}

We show qualitative results in \ref{fig:iclight} and \ref{fig:icgrid}. IC-Light is temporally inconsisent and is not able to produce precise lighting from a given direction. Additionally, because it only takes in a background image, it cannot be used to relight a subject given a full 360 HDRI map. Regardless of the prompt chosen, or the background chosen, IC-Light fails to add reflections to the subject's eyes - resulting in a cloudy, lifeless appearance. By contrast, our method accurately reproduces specularities in high fidelity. These results showcase the importance of our subject-specific formulation.


\subsection{Additional Ablation Studies}

\subsubsection{Ablation on SH degree}

One hyper-parameters in our method is the SH degree used for encoding the light direction into text embedding. We experiment with different SH degrees on one subject and plot the PSNR and SSIM metrics in Fig.~\ref{fig:ablation_sh} under SH degree $= 2, 3, 5, 7$. It can be see that when degree equals to 2, our method struggles to reason on the light direction and produce high quality results, which is likely due to the low frequency band of the light condition. When degree is larger than $2$, the model achieves much better performance and achieves the best when the degree is $3$, which is the final settings we choose in all of other experiments. This specific value roughly aligns with the angular sampling rate of all the possible lighting directions in the dataset. Therefore, choosing a proper SH degree is important for training a good personalized relighting model. 

\subsubsection{Ablation Studies on Scalable Dynamic 3DGS}
\label{sec:ab_dynamic_gs}
\begin{figure*}[t]
\centering
\renewcommand{\arraystretch}{0}
\begin{tabular}{cc}
\raisebox{3.5em}{\rotatebox{90}{GT}} & \includegraphics[width=0.75\linewidth]{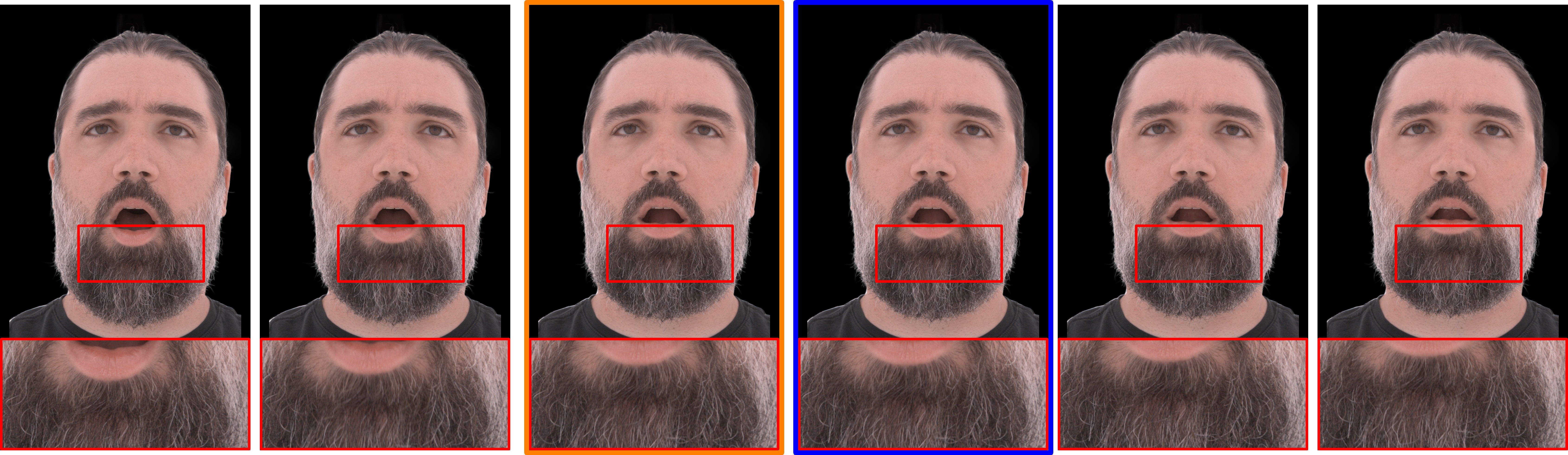} \\
\raisebox{3.5em}{\rotatebox{90}{(a)}} & \includegraphics[width=0.75\linewidth]{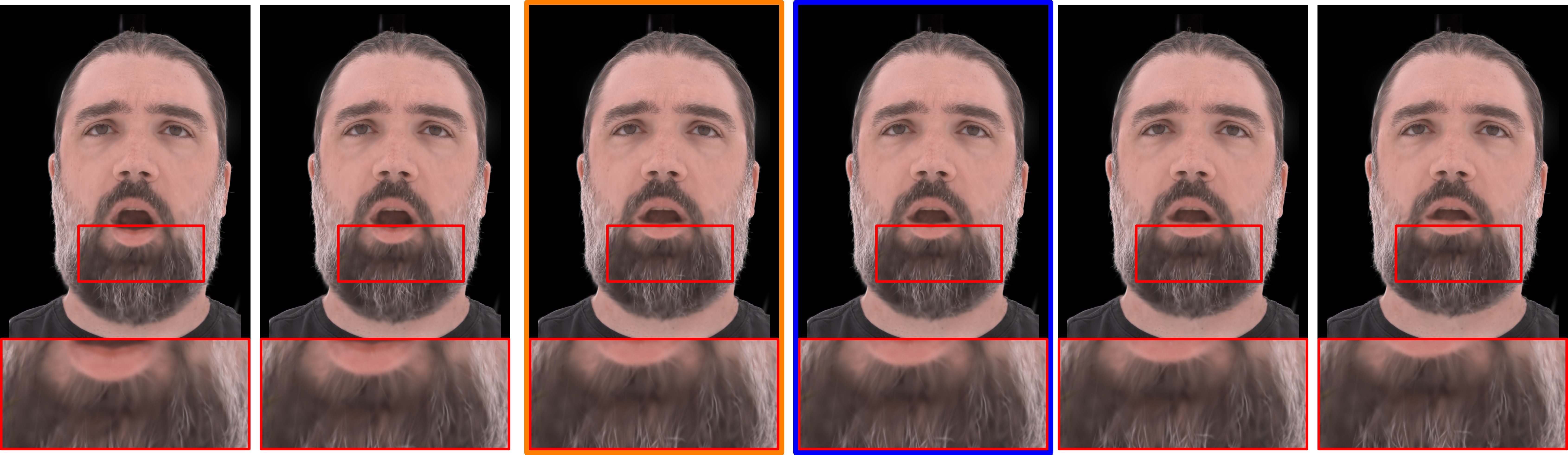} \\
\raisebox{3.5em}{\rotatebox{90}{(b)}} & \includegraphics[width=0.75\linewidth]{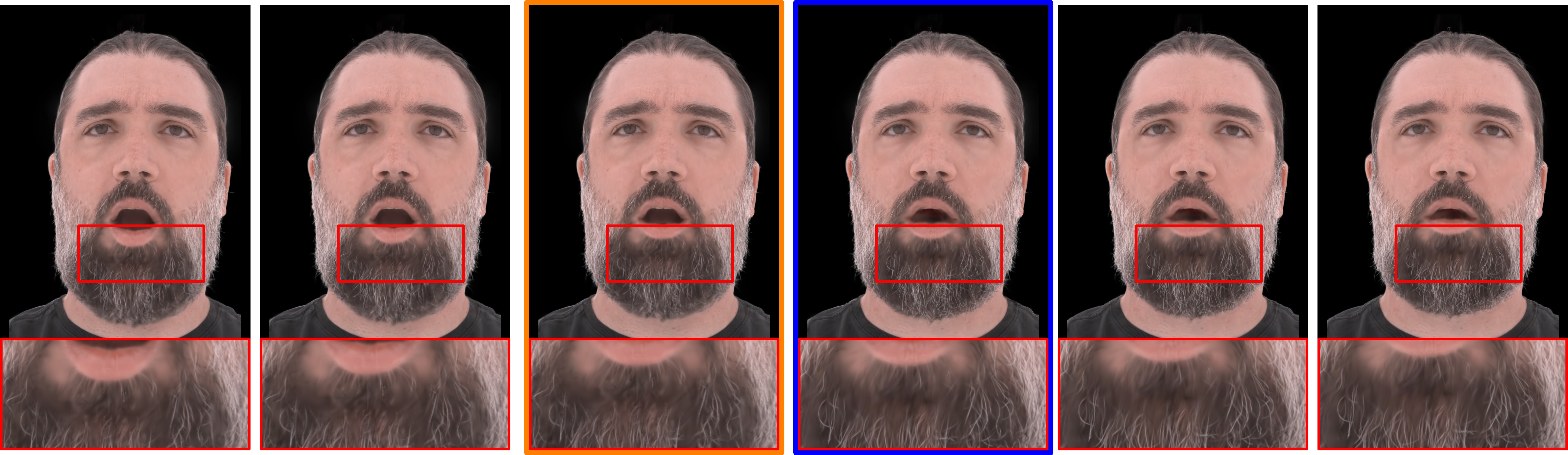} \\
\raisebox{3.5em}{\rotatebox{90}{(c)}} & \includegraphics[width=0.75\linewidth]{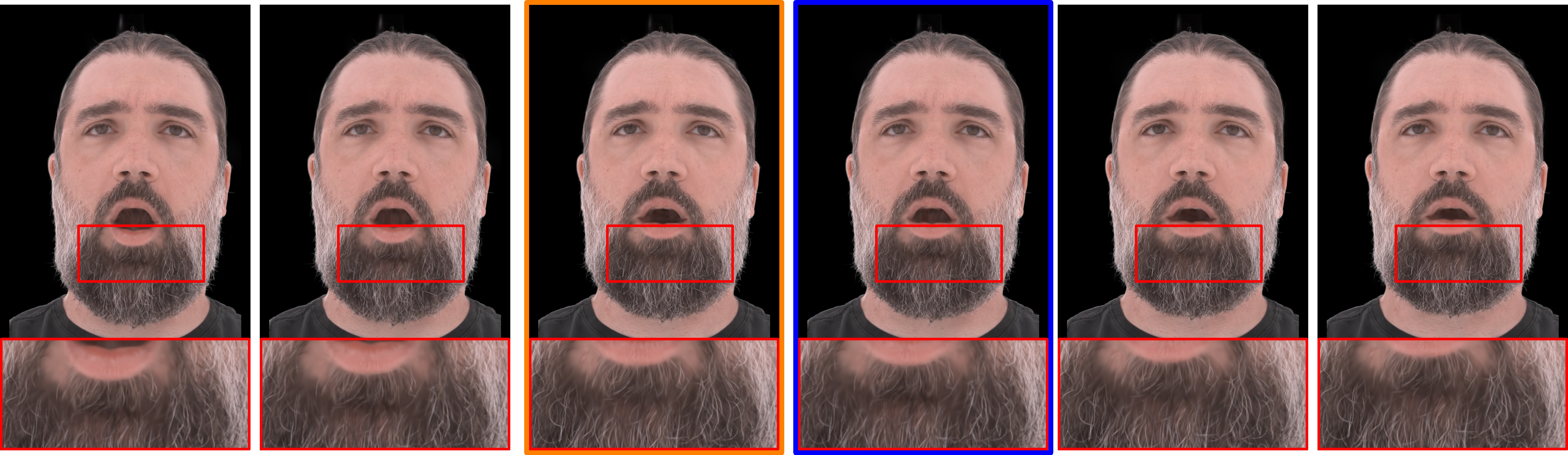} \\
\raisebox{3.5em}{\rotatebox{90}{(d)}} & \includegraphics[width=0.75\linewidth]{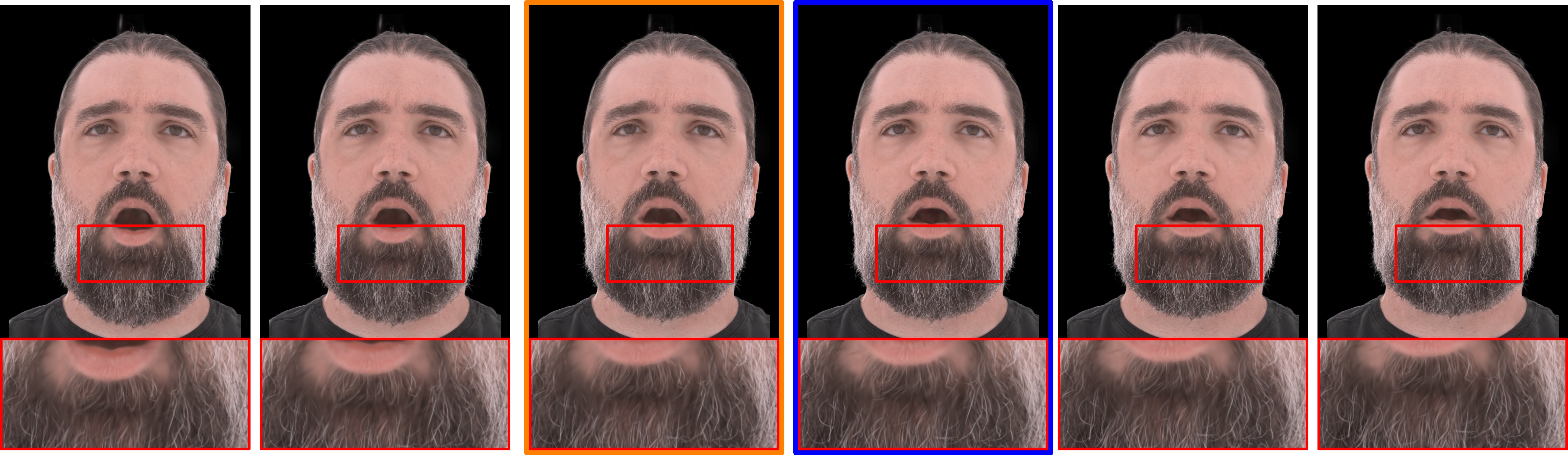} 
\end{tabular}
\caption{Visual comparison on different dynamic 3DGS models: (a) baseline model~\cite{jung2023_deformable_gs}, (b) partitions long sequences into segments for separate training, (c) uses our K-frame initialization without deformation offset regularization, and (d) combines K-frame initialization with deformation offset regularization. Orange and blue frames denote transition points overlapped by two segments in (b), (c), and (d). The orange frame is learned by the previous segment model, and the blue frame comes from the next segment model, indicating better cross-segment consistency with greater similarity. In, GT and (a), they are duplicates and identical.}
\label{fig:ablation_temporal}
\end{figure*}

We show the visual results for the ablation study on the proposed dynamic 3DGS in Fig.~\ref{fig:ablation_temporal}.

%% file: main.bbl

\begin{thebibliography}{80}


\ifx \showCODEN    \undefined \def \showCODEN     #1{\unskip}     \fi
\ifx \showDOI      \undefined \def \showDOI       #1{#1}\fi
\ifx \showISBNx    \undefined \def \showISBNx     #1{\unskip}     \fi
\ifx \showISBNxiii \undefined \def \showISBNxiii  #1{\unskip}     \fi
\ifx \showISSN     \undefined \def \showISSN      #1{\unskip}     \fi
\ifx \showLCCN     \undefined \def \showLCCN      #1{\unskip}     \fi
\ifx \shownote     \undefined \def \shownote      #1{#1}          \fi
\ifx \showarticletitle \undefined \def \showarticletitle #1{#1}   \fi
\ifx \showURL      \undefined \def \showURL       {\relax}        \fi
\providecommand\bibfield[2]{#2}
\providecommand\bibinfo[2]{#2}
\providecommand\natexlab[1]{#1}
\providecommand\showeprint[2][]{arXiv:#2}

\bibitem[Andersson et~al\mbox{.}(2020)]%
        {andersson2020flip}
\bibfield{author}{\bibinfo{person}{Pontus Andersson}, \bibinfo{person}{Jim Nilsson}, \bibinfo{person}{Tomas Akenine-M{\"o}ller}, \bibinfo{person}{Magnus Oskarsson}, \bibinfo{person}{Kalle {\AA}str{\"o}m}, {and} \bibinfo{person}{Mark~D Fairchild}.} \bibinfo{year}{2020}\natexlab{}.
\newblock \showarticletitle{FLIP: A Difference Evaluator for Alternating Images.}
\newblock \bibinfo{journal}{\emph{Proc. ACM Comput. Graph. Interact. Tech.}} \bibinfo{volume}{3}, \bibinfo{number}{2} (\bibinfo{year}{2020}), \bibinfo{pages}{15--1}.
\newblock


\bibitem[Barron and Malik(2014)]%
        {barron2014shape}
\bibfield{author}{\bibinfo{person}{Jonathan~T Barron} {and} \bibinfo{person}{Jitendra Malik}.} \bibinfo{year}{2014}\natexlab{}.
\newblock \showarticletitle{Shape, illumination, and reflectance from shading}.
\newblock \bibinfo{journal}{\emph{IEEE transactions on pattern analysis and machine intelligence}} \bibinfo{volume}{37}, \bibinfo{number}{8} (\bibinfo{year}{2014}), \bibinfo{pages}{1670--1687}.
\newblock


\bibitem[Boss et~al\mbox{.}(2021)]%
        {boss2021_nerd}
\bibfield{author}{\bibinfo{person}{Mark Boss}, \bibinfo{person}{Raphael Braun}, \bibinfo{person}{Varun Jampani}, \bibinfo{person}{Jonathan~T. Barron}, \bibinfo{person}{Ce Liu}, {and} \bibinfo{person}{Hendrik P.~A. Lensch}.} \bibinfo{year}{2021}\natexlab{}.
\newblock \bibinfo{title}{NeRD: Neural Reflectance Decomposition from Image Collections}.
\newblock
\newblock
\showeprint[arxiv]{2012.03918}~[cs.CV]


\bibitem[Brooks et~al\mbox{.}(2023)]%
        {brooks2023instructpix2pix}
\bibfield{author}{\bibinfo{person}{Tim Brooks}, \bibinfo{person}{Aleksander Holynski}, {and} \bibinfo{person}{Alexei~A Efros}.} \bibinfo{year}{2023}\natexlab{}.
\newblock \showarticletitle{Instructpix2pix: Learning to follow image editing instructions}. In \bibinfo{booktitle}{\emph{Proceedings of the IEEE/CVF Conference on Computer Vision and Pattern Recognition}}. \bibinfo{pages}{18392--18402}.
\newblock


\bibitem[Debevec and LeGendre(2022)]%
        {debevec2022hdrlightingdilationdynamic}
\bibfield{author}{\bibinfo{person}{Paul Debevec} {and} \bibinfo{person}{Chloe LeGendre}.} \bibinfo{year}{2022}\natexlab{}.
\newblock \bibinfo{title}{HDR Lighting Dilation for Dynamic Range Reduction on Virtual Production Stages}.
\newblock
\newblock
\showeprint[arxiv]{2205.07873}~[cs.GR]
\urldef\tempurl%
\url{https://arxiv.org/abs/2205.07873}
\showURL{%
\tempurl}


\bibitem[Debevec et~al\mbox{.}(2000)]%
        {debevec2000_acquiring_reflectance_field}
\bibfield{author}{\bibinfo{person}{Paul~E. Debevec}, \bibinfo{person}{Tim Hawkins}, \bibinfo{person}{Chris Tchou}, \bibinfo{person}{Haarm{-}Pieter Duiker}, \bibinfo{person}{Westley Sarokin}, {and} \bibinfo{person}{Mark Sagar}.} \bibinfo{year}{2000}\natexlab{}.
\newblock \showarticletitle{Acquiring the reflectance field of a human face}. In \bibinfo{booktitle}{\emph{Proceedings of the 27th Annual Conference on Computer Graphics and Interactive Techniques, {SIGGRAPH} 2000, New Orleans, LA, USA, July 23-28, 2000}}, \bibfield{editor}{\bibinfo{person}{Judith~R. Brown} {and} \bibinfo{person}{Kurt Akeley}} (Eds.). \bibinfo{publisher}{{ACM}}, \bibinfo{pages}{145--156}.
\newblock
\urldef\tempurl%
\url{https://doi.org/10.1145/344779.344855}
\showDOI{\tempurl}


\bibitem[Ding et~al\mbox{.}(2023)]%
        {ding2023diffusionrig}
\bibfield{author}{\bibinfo{person}{Zheng Ding}, \bibinfo{person}{Xuaner Zhang}, \bibinfo{person}{Zhihao Xia}, \bibinfo{person}{Lars Jebe}, \bibinfo{person}{Zhuowen Tu}, {and} \bibinfo{person}{Xiuming Zhang}.} \bibinfo{year}{2023}\natexlab{}.
\newblock \showarticletitle{Diffusionrig: Learning personalized priors for facial appearance editing}. In \bibinfo{booktitle}{\emph{Proceedings of the IEEE/CVF Conference on Computer Vision and Pattern Recognition}}. \bibinfo{pages}{12736--12746}.
\newblock


\bibitem[Einarsson et~al\mbox{.}(2006)]%
        {Einarsson:2006:RHL}
\bibfield{author}{\bibinfo{person}{Per Einarsson}, \bibinfo{person}{Charles-Felix Chabert}, \bibinfo{person}{Andrew Jones}, \bibinfo{person}{Wan-Chun Ma}, \bibinfo{person}{Bruce Lamond}, \bibinfo{person}{Tim Hawkins}, \bibinfo{person}{Mark Bolas}, \bibinfo{person}{Sebastian Sylwan}, {and} \bibinfo{person}{Paul Debevec}.} \bibinfo{year}{2006}\natexlab{}.
\newblock \showarticletitle{Relighting human locomotion with flowed reflectance fields}. In \bibinfo{booktitle}{\emph{Proceedings of the 17th Eurographics Conference on Rendering Techniques}} (Nicosia, Cyprus) \emph{(\bibinfo{series}{EGSR '06})}. \bibinfo{publisher}{Eurographics Association}, \bibinfo{address}{Goslar, DEU}, \bibinfo{pages}{183–194}.
\newblock
\showISBNx{3905673355}


\bibitem[Face(2024)]%
        {HuggingFace_Diffusers}
\bibfield{author}{\bibinfo{person}{Hugging Face}.} \bibinfo{year}{2024}\natexlab{}.
\newblock \bibinfo{title}{Diffusers Documentation}.
\newblock
\newblock
\urldef\tempurl%
\url{https://huggingface.co/docs/diffusers/index}
\showURL{%
\tempurl}
\newblock
\shownote{Accessed: 2024-05-19}.


\bibitem[Gao et~al\mbox{.}(2023)]%
        {gao2023_relightable_3d_gaussians}
\bibfield{author}{\bibinfo{person}{Jian Gao}, \bibinfo{person}{Chun Gu}, \bibinfo{person}{Youtian Lin}, \bibinfo{person}{Hao Zhu}, \bibinfo{person}{Xun Cao}, \bibinfo{person}{Li Zhang}, {and} \bibinfo{person}{Yao Yao}.} \bibinfo{year}{2023}\natexlab{}.
\newblock \showarticletitle{Relightable 3D Gaussian: Real-time Point Cloud Relighting with {BRDF} Decomposition and Ray Tracing}.
\newblock \bibinfo{journal}{\emph{CoRR}}  \bibinfo{volume}{abs/2311.16043} (\bibinfo{year}{2023}).
\newblock
\urldef\tempurl%
\url{https://doi.org/10.48550/ARXIV.2311.16043}
\showDOI{\tempurl}
\showeprint[arXiv]{2311.16043}


\bibitem[Goesele et~al\mbox{.}(2004)]%
        {Goesele:2004:DISCO}
\bibfield{author}{\bibinfo{person}{Michael Goesele}, \bibinfo{person}{Hendrik P.~A. Lensch}, \bibinfo{person}{Jochen Lang}, \bibinfo{person}{Christian Fuchs}, {and} \bibinfo{person}{Hans-Peter Seidel}.} \bibinfo{year}{2004}\natexlab{}.
\newblock \showarticletitle{DISCO: acquisition of translucent objects}.
\newblock \bibinfo{journal}{\emph{ACM Trans. Graph.}} \bibinfo{volume}{23}, \bibinfo{number}{3} (\bibinfo{date}{aug} \bibinfo{year}{2004}), \bibinfo{pages}{835–844}.
\newblock
\showISSN{0730-0301}
\urldef\tempurl%
\url{https://doi.org/10.1145/1015706.1015807}
\showDOI{\tempurl}


\bibitem[Guo et~al\mbox{.}(2019)]%
        {Guo:therelightables:2019}
\bibfield{author}{\bibinfo{person}{Kaiwen Guo}, \bibinfo{person}{Peter Lincoln}, \bibinfo{person}{Philip Davidson}, \bibinfo{person}{Jay Busch}, \bibinfo{person}{Xueming Yu}, \bibinfo{person}{Matt Whalen}, \bibinfo{person}{Geoff Harvey}, \bibinfo{person}{Sergio Orts-Escolano}, \bibinfo{person}{Rohit Pandey}, \bibinfo{person}{Jason Dourgarian}, \bibinfo{person}{Danhang Tang}, \bibinfo{person}{Anastasia Tkach}, \bibinfo{person}{Adarsh Kowdle}, \bibinfo{person}{Emily Cooper}, \bibinfo{person}{Mingsong Dou}, \bibinfo{person}{Sean Fanello}, \bibinfo{person}{Graham Fyffe}, \bibinfo{person}{Christoph Rhemann}, \bibinfo{person}{Jonathan Taylor}, \bibinfo{person}{Paul Debevec}, {and} \bibinfo{person}{Shahram Izadi}.} \bibinfo{year}{2019}\natexlab{}.
\newblock \showarticletitle{The relightables: volumetric performance capture of humans with realistic relighting}.
\newblock \bibinfo{journal}{\emph{ACM Trans. Graph.}} \bibinfo{volume}{38}, \bibinfo{number}{6}, Article \bibinfo{articleno}{217} (\bibinfo{date}{nov} \bibinfo{year}{2019}), \bibinfo{numpages}{19}~pages.
\newblock
\showISSN{0730-0301}
\urldef\tempurl%
\url{https://doi.org/10.1145/3355089.3356571}
\showDOI{\tempurl}


\bibitem[Guttenberg(2023)]%
        {offsetnoise}
\bibfield{author}{\bibinfo{person}{Nicholas Guttenberg}.} \bibinfo{year}{2023}\natexlab{}.
\newblock \bibinfo{title}{Diffusion with Offset Noise}.
\newblock \bibinfo{howpublished}{\url{https://www.crosslabs.org/blog/diffusion-with-offset-noise}}.
\newblock
\newblock
\shownote{Accessed: 2024-05-15}.


\bibitem[He et~al\mbox{.}(2016)]%
        {he2015deepres}
\bibfield{author}{\bibinfo{person}{Kaiming He}, \bibinfo{person}{Xiangyu Zhang}, \bibinfo{person}{Shaoqing Ren}, {and} \bibinfo{person}{Jian Sun}.} \bibinfo{year}{2016}\natexlab{}.
\newblock \showarticletitle{Deep Residual Learning for Image Recognition}. In \bibinfo{booktitle}{\emph{2016 IEEE Conference on Computer Vision and Pattern Recognition (CVPR)}}. \bibinfo{pages}{770--778}.
\newblock
\urldef\tempurl%
\url{https://doi.org/10.1109/CVPR.2016.90}
\showDOI{\tempurl}


\bibitem[Jamri{\v{s}}ka et~al\mbox{.}(2019)]%
        {jamrivska2019stylizing}
\bibfield{author}{\bibinfo{person}{Ond{\v{r}}ej Jamri{\v{s}}ka}, \bibinfo{person}{{\v{S}}{\'a}rka Sochorov{\'a}}, \bibinfo{person}{Ond{\v{r}}ej Texler}, \bibinfo{person}{Michal Luk{\'a}{\v{c}}}, \bibinfo{person}{Jakub Fi{\v{s}}er}, \bibinfo{person}{Jingwan Lu}, \bibinfo{person}{Eli Shechtman}, {and} \bibinfo{person}{Daniel S{\`y}kora}.} \bibinfo{year}{2019}\natexlab{}.
\newblock \showarticletitle{Stylizing video by example}.
\newblock \bibinfo{journal}{\emph{ACM Transactions on Graphics (TOG)}} \bibinfo{volume}{38}, \bibinfo{number}{4} (\bibinfo{year}{2019}), \bibinfo{pages}{1--11}.
\newblock


\bibitem[Jung et~al\mbox{.}(2023)]%
        {jung2023_deformable_gs}
\bibfield{author}{\bibinfo{person}{HyunJun Jung}, \bibinfo{person}{Nikolas Brasch}, \bibinfo{person}{Jifei Song}, \bibinfo{person}{Eduardo Perez-Pellitero}, \bibinfo{person}{Yiren Zhou}, \bibinfo{person}{Zhihao Li}, \bibinfo{person}{Nassir Navab}, {and} \bibinfo{person}{Benjamin Busam}.} \bibinfo{year}{2023}\natexlab{}.
\newblock \bibinfo{title}{Deformable 3D Gaussian Splatting for Animatable Human Avatars}.
\newblock
\newblock
\showeprint[arxiv]{2312.15059}~[cs.CV]


\bibitem[Karras et~al\mbox{.}(2022)]%
        {karras2022elucidating}
\bibfield{author}{\bibinfo{person}{Tero Karras}, \bibinfo{person}{Miika Aittala}, \bibinfo{person}{Timo Aila}, {and} \bibinfo{person}{Samuli Laine}.} \bibinfo{year}{2022}\natexlab{}.
\newblock \showarticletitle{Elucidating the design space of diffusion-based generative models}.
\newblock \bibinfo{journal}{\emph{Advances in Neural Information Processing Systems}}  \bibinfo{volume}{35} (\bibinfo{year}{2022}), \bibinfo{pages}{26565--26577}.
\newblock


\bibitem[Karras et~al\mbox{.}(2020)]%
        {karras2020styleGANQuality}
\bibfield{author}{\bibinfo{person}{Tero Karras}, \bibinfo{person}{Samuli Laine}, \bibinfo{person}{Miika Aittala}, \bibinfo{person}{Janne Hellsten}, \bibinfo{person}{Jaakko Lehtinen}, {and} \bibinfo{person}{Timo Aila}.} \bibinfo{year}{2020}\natexlab{}.
\newblock \showarticletitle{Analyzing and Improving the Image Quality of StyleGAN}. In \bibinfo{booktitle}{\emph{2020 IEEE/CVF Conference on Computer Vision and Pattern Recognition (CVPR)}}. \bibinfo{pages}{8107--8116}.
\newblock
\urldef\tempurl%
\url{https://doi.org/10.1109/CVPR42600.2020.00813}
\showDOI{\tempurl}


\bibitem[Ke et~al\mbox{.}(2023)]%
        {ke2023repurposing}
\bibfield{author}{\bibinfo{person}{Bingxin Ke}, \bibinfo{person}{Anton Obukhov}, \bibinfo{person}{Shengyu Huang}, \bibinfo{person}{Nando Metzger}, \bibinfo{person}{Rodrigo~Caye Daudt}, {and} \bibinfo{person}{Konrad Schindler}.} \bibinfo{year}{2023}\natexlab{}.
\newblock \showarticletitle{Repurposing diffusion-based image generators for monocular depth estimation}.
\newblock \bibinfo{journal}{\emph{arXiv preprint arXiv:2312.02145}} (\bibinfo{year}{2023}).
\newblock


\bibitem[Kerbl et~al\mbox{.}(2023)]%
        {kerbl2023_3d_gs}
\bibfield{author}{\bibinfo{person}{Bernhard Kerbl}, \bibinfo{person}{Georgios Kopanas}, \bibinfo{person}{Thomas Leimk{\"{u}}hler}, {and} \bibinfo{person}{George Drettakis}.} \bibinfo{year}{2023}\natexlab{}.
\newblock \showarticletitle{3D Gaussian Splatting for Real-Time Radiance Field Rendering}.
\newblock \bibinfo{journal}{\emph{{ACM} Trans. Graph.}} \bibinfo{volume}{42}, \bibinfo{number}{4} (\bibinfo{year}{2023}), \bibinfo{pages}{139:1--139:14}.
\newblock
\urldef\tempurl%
\url{https://doi.org/10.1145/3592433}
\showDOI{\tempurl}


\bibitem[Kim et~al\mbox{.}(2024)]%
        {kim2024switchlightcodesignphysicsdrivenarchitecture}
\bibfield{author}{\bibinfo{person}{Hoon Kim}, \bibinfo{person}{Minje Jang}, \bibinfo{person}{Wonjun Yoon}, \bibinfo{person}{Jisoo Lee}, \bibinfo{person}{Donghyun Na}, {and} \bibinfo{person}{Sanghyun Woo}.} \bibinfo{year}{2024}\natexlab{}.
\newblock \bibinfo{title}{SwitchLight: Co-design of Physics-driven Architecture and Pre-training Framework for Human Portrait Relighting}.
\newblock
\newblock
\showeprint[arxiv]{2402.18848}~[cs.CV]
\urldef\tempurl%
\url{https://arxiv.org/abs/2402.18848}
\showURL{%
\tempurl}


\bibitem[Le and Kakadiaris(2019)]%
        {le2019illumination}
\bibfield{author}{\bibinfo{person}{Ha~A Le} {and} \bibinfo{person}{Ioannis~A Kakadiaris}.} \bibinfo{year}{2019}\natexlab{}.
\newblock \showarticletitle{Illumination-invariant face recognition with deep relit face images}. In \bibinfo{booktitle}{\emph{2019 IEEE Winter Conference on Applications of Computer Vision (WACV)}}. IEEE, \bibinfo{pages}{2146--2155}.
\newblock


\bibitem[LeGendre et~al\mbox{.}(2020)]%
        {LeGendre_olat}
\bibfield{author}{\bibinfo{person}{Chloe LeGendre}, \bibinfo{person}{Wan{-}Chun Ma}, \bibinfo{person}{Rohit Pandey}, \bibinfo{person}{Sean~Ryan Fanello}, \bibinfo{person}{Christoph Rhemann}, \bibinfo{person}{Jason Dourgarian}, \bibinfo{person}{Jay Busch}, {and} \bibinfo{person}{Paul~E. Debevec}.} \bibinfo{year}{2020}\natexlab{}.
\newblock \showarticletitle{Learning Illumination from Diverse Portraits}. In \bibinfo{booktitle}{\emph{{SA} '20: {SIGGRAPH} Asia 2020 Technical Communications, Virtual Event, Republic of Korea, December, 2020}}. \bibinfo{publisher}{{ACM}}, \bibinfo{pages}{7:1--7:4}.
\newblock
\urldef\tempurl%
\url{https://doi.org/10.1145/3410700.3425432}
\showDOI{\tempurl}


\bibitem[Li et~al\mbox{.}(2018)]%
        {li2018closed}
\bibfield{author}{\bibinfo{person}{Yijun Li}, \bibinfo{person}{Ming-Yu Liu}, \bibinfo{person}{Xueting Li}, \bibinfo{person}{Ming-Hsuan Yang}, {and} \bibinfo{person}{Jan Kautz}.} \bibinfo{year}{2018}\natexlab{}.
\newblock \showarticletitle{A closed-form solution to photorealistic image stylization}. In \bibinfo{booktitle}{\emph{Proceedings of the European conference on computer vision (ECCV)}}. \bibinfo{pages}{453--468}.
\newblock


\bibitem[Li et~al\mbox{.}(2023a)]%
        {li2023spacetime}
\bibfield{author}{\bibinfo{person}{Zhan Li}, \bibinfo{person}{Zhang Chen}, \bibinfo{person}{Zhong Li}, {and} \bibinfo{person}{Yi Xu}.} \bibinfo{year}{2023}\natexlab{a}.
\newblock \showarticletitle{Spacetime gaussian feature splatting for real-time dynamic view synthesis}.
\newblock \bibinfo{journal}{\emph{arXiv preprint arXiv:2312.16812}} (\bibinfo{year}{2023}).
\newblock


\bibitem[Li et~al\mbox{.}(2023b)]%
        {li2023_relitneulf}
\bibfield{author}{\bibinfo{person}{Zhong Li}, \bibinfo{person}{Liangchen Song}, \bibinfo{person}{Zhang Chen}, \bibinfo{person}{Xiangyu Du}, \bibinfo{person}{Lele Chen}, \bibinfo{person}{Junsong Yuan}, {and} \bibinfo{person}{Yi Xu}.} \bibinfo{year}{2023}\natexlab{b}.
\newblock \showarticletitle{Relit-NeuLF: Efficient Relighting and Novel View Synthesis via Neural 4D Light Field}. In \bibinfo{booktitle}{\emph{Proceedings of the 31st {ACM} International Conference on Multimedia, {MM} 2023, Ottawa, ON, Canada, 29 October 2023- 3 November 2023}}, \bibfield{editor}{\bibinfo{person}{Abdulmotaleb El{-}Saddik}, \bibinfo{person}{Tao Mei}, \bibinfo{person}{Rita Cucchiara}, \bibinfo{person}{Marco Bertini}, \bibinfo{person}{Diana Patricia~Tobon Vallejo}, \bibinfo{person}{Pradeep~K. Atrey}, {and} \bibinfo{person}{M.~Shamim Hossain}} (Eds.). \bibinfo{publisher}{{ACM}}, \bibinfo{pages}{7007--7016}.
\newblock
\urldef\tempurl%
\url{https://doi.org/10.1145/3581783.3612160}
\showDOI{\tempurl}


\bibitem[Liu et~al\mbox{.}(2023)]%
        {liu2023_nero}
\bibfield{author}{\bibinfo{person}{Yuan Liu}, \bibinfo{person}{Peng Wang}, \bibinfo{person}{Cheng Lin}, \bibinfo{person}{Xiaoxiao Long}, \bibinfo{person}{Jiepeng Wang}, \bibinfo{person}{Lingjie Liu}, \bibinfo{person}{Taku Komura}, {and} \bibinfo{person}{Wenping Wang}.} \bibinfo{year}{2023}\natexlab{}.
\newblock \showarticletitle{NeRO: Neural Geometry and {BRDF} Reconstruction of Reflective Objects from Multiview Images}.
\newblock \bibinfo{journal}{\emph{{ACM} Trans. Graph.}} \bibinfo{volume}{42}, \bibinfo{number}{4} (\bibinfo{year}{2023}), \bibinfo{pages}{114:1--114:22}.
\newblock
\urldef\tempurl%
\url{https://doi.org/10.1145/3592134}
\showDOI{\tempurl}


\bibitem[Luan et~al\mbox{.}(2017)]%
        {luan2017deep}
\bibfield{author}{\bibinfo{person}{Fujun Luan}, \bibinfo{person}{Sylvain Paris}, \bibinfo{person}{Eli Shechtman}, {and} \bibinfo{person}{Kavita Bala}.} \bibinfo{year}{2017}\natexlab{}.
\newblock \showarticletitle{Deep photo style transfer}. In \bibinfo{booktitle}{\emph{Proceedings of the IEEE conference on computer vision and pattern recognition}}. \bibinfo{pages}{4990--4998}.
\newblock


\bibitem[Luiten et~al\mbox{.}(2023)]%
        {luiten2023_dynamic_gs}
\bibfield{author}{\bibinfo{person}{Jonathon Luiten}, \bibinfo{person}{Georgios Kopanas}, \bibinfo{person}{Bastian Leibe}, {and} \bibinfo{person}{Deva Ramanan}.} \bibinfo{year}{2023}\natexlab{}.
\newblock \showarticletitle{Dynamic 3D Gaussians: Tracking by Persistent Dynamic View Synthesis}.
\newblock \bibinfo{journal}{\emph{CoRR}}  \bibinfo{volume}{abs/2308.09713} (\bibinfo{year}{2023}).
\newblock
\urldef\tempurl%
\url{https://doi.org/10.48550/ARXIV.2308.09713}
\showDOI{\tempurl}
\showeprint[arXiv]{2308.09713}


\bibitem[Ma et~al\mbox{.}(2007)]%
        {ma2007rapid}
\bibfield{author}{\bibinfo{person}{Wan-Chun Ma}, \bibinfo{person}{Tim Hawkins}, \bibinfo{person}{Pieter Peers}, \bibinfo{person}{Charles-Felix Chabert}, \bibinfo{person}{Malte Weiss}, \bibinfo{person}{Paul~E Debevec}, {et~al\mbox{.}}} \bibinfo{year}{2007}\natexlab{}.
\newblock \showarticletitle{Rapid Acquisition of Specular and Diffuse Normal Maps from Polarized Spherical Gradient Illumination.}
\newblock \bibinfo{journal}{\emph{Rendering Techniques}} \bibinfo{volume}{9}, \bibinfo{number}{10} (\bibinfo{year}{2007}), \bibinfo{pages}{2}.
\newblock


\bibitem[Mei et~al\mbox{.}(2024)]%
        {mei2024holo}
\bibfield{author}{\bibinfo{person}{Yiqun Mei}, \bibinfo{person}{Yu Zeng}, \bibinfo{person}{He Zhang}, \bibinfo{person}{Zhixin Shu}, \bibinfo{person}{Xuaner Zhang}, \bibinfo{person}{Sai Bi}, \bibinfo{person}{Jianming Zhang}, \bibinfo{person}{HyunJoon Jung}, {and} \bibinfo{person}{Vishal~M Patel}.} \bibinfo{year}{2024}\natexlab{}.
\newblock \showarticletitle{Holo-Relighting: Controllable Volumetric Portrait Relighting from a Single Image}.
\newblock \bibinfo{journal}{\emph{arXiv preprint arXiv:2403.09632}} (\bibinfo{year}{2024}).
\newblock


\bibitem[Meka et~al\mbox{.}(2019)]%
        {meka2019deep}
\bibfield{author}{\bibinfo{person}{Abhimitra Meka}, \bibinfo{person}{Christian Haene}, \bibinfo{person}{Rohit Pandey}, \bibinfo{person}{Michael Zollh{\"o}fer}, \bibinfo{person}{Sean Fanello}, \bibinfo{person}{Graham Fyffe}, \bibinfo{person}{Adarsh Kowdle}, \bibinfo{person}{Xueming Yu}, \bibinfo{person}{Jay Busch}, \bibinfo{person}{Jason Dourgarian}, {et~al\mbox{.}}} \bibinfo{year}{2019}\natexlab{}.
\newblock \showarticletitle{Deep reflectance fields: high-quality facial reflectance field inference from color gradient illumination}.
\newblock \bibinfo{journal}{\emph{ACM Transactions on Graphics (TOG)}} \bibinfo{volume}{38}, \bibinfo{number}{4} (\bibinfo{year}{2019}), \bibinfo{pages}{1--12}.
\newblock


\bibitem[Meka et~al\mbox{.}(2020)]%
        {meka2020_deep_relightable_textures}
\bibfield{author}{\bibinfo{person}{Abhimitra Meka}, \bibinfo{person}{Rohit Pandey}, \bibinfo{person}{Christian H{\"{a}}ne}, \bibinfo{person}{Sergio Orts{-}Escolano}, \bibinfo{person}{Peter Barnum}, \bibinfo{person}{Philip~L. Davidson}, \bibinfo{person}{Daniel Erickson}, \bibinfo{person}{Yinda Zhang}, \bibinfo{person}{Jonathan Taylor}, \bibinfo{person}{Sofien Bouaziz}, \bibinfo{person}{Chloe LeGendre}, \bibinfo{person}{Wan{-}Chun Ma}, \bibinfo{person}{Ryan~S. Overbeck}, \bibinfo{person}{Thabo Beeler}, \bibinfo{person}{Paul~E. Debevec}, \bibinfo{person}{Shahram Izadi}, \bibinfo{person}{Christian Theobalt}, \bibinfo{person}{Christoph Rhemann}, {and} \bibinfo{person}{Sean~Ryan Fanello}.} \bibinfo{year}{2020}\natexlab{}.
\newblock \showarticletitle{Deep relightable textures: volumetric performance capture with neural rendering}.
\newblock \bibinfo{journal}{\emph{{ACM} Trans. Graph.}} \bibinfo{volume}{39}, \bibinfo{number}{6} (\bibinfo{year}{2020}), \bibinfo{pages}{259:1--259:21}.
\newblock
\urldef\tempurl%
\url{https://doi.org/10.1145/3414685.3417814}
\showDOI{\tempurl}


\bibitem[Meng et~al\mbox{.}(2021)]%
        {meng2021sdedit}
\bibfield{author}{\bibinfo{person}{Chenlin Meng}, \bibinfo{person}{Yutong He}, \bibinfo{person}{Yang Song}, \bibinfo{person}{Jiaming Song}, \bibinfo{person}{Jiajun Wu}, \bibinfo{person}{Jun-Yan Zhu}, {and} \bibinfo{person}{Stefano Ermon}.} \bibinfo{year}{2021}\natexlab{}.
\newblock \showarticletitle{Sdedit: Guided image synthesis and editing with stochastic differential equations}.
\newblock \bibinfo{journal}{\emph{arXiv preprint arXiv:2108.01073}} (\bibinfo{year}{2021}).
\newblock


\bibitem[Mildenhall et~al\mbox{.}(2020)]%
        {mildenhall2020_nerf}
\bibfield{author}{\bibinfo{person}{Ben Mildenhall}, \bibinfo{person}{Pratul~P. Srinivasan}, \bibinfo{person}{Matthew Tancik}, \bibinfo{person}{Jonathan~T. Barron}, \bibinfo{person}{Ravi Ramamoorthi}, {and} \bibinfo{person}{Ren Ng}.} \bibinfo{year}{2020}\natexlab{}.
\newblock \bibinfo{title}{NeRF: Representing Scenes as Neural Radiance Fields for View Synthesis}.
\newblock
\newblock
\showeprint[arxiv]{2003.08934}~[cs.CV]


\bibitem[Nimeroff et~al\mbox{.}(1995)]%
        {nimeroff1995efficient}
\bibfield{author}{\bibinfo{person}{Jeffry~S Nimeroff}, \bibinfo{person}{Eero Simoncelli}, {and} \bibinfo{person}{Julie Dorsey}.} \bibinfo{year}{1995}\natexlab{}.
\newblock \showarticletitle{Efficient re-rendering of naturally illuminated environments}. In \bibinfo{booktitle}{\emph{Photorealistic Rendering Techniques}}. Springer, \bibinfo{pages}{373--388}.
\newblock


\bibitem[Pandey et~al\mbox{.}(2021)]%
        {pandey2021_total_relighting}
\bibfield{author}{\bibinfo{person}{Rohit Pandey}, \bibinfo{person}{Sergio Orts{-}Escolano}, \bibinfo{person}{Chloe LeGendre}, \bibinfo{person}{Christian H{\"{a}}ne}, \bibinfo{person}{Sofien Bouaziz}, \bibinfo{person}{Christoph Rhemann}, \bibinfo{person}{Paul~E. Debevec}, {and} \bibinfo{person}{Sean~Ryan Fanello}.} \bibinfo{year}{2021}\natexlab{}.
\newblock \showarticletitle{Total relighting: learning to relight portraits for background replacement}.
\newblock \bibinfo{journal}{\emph{{ACM} Trans. Graph.}} \bibinfo{volume}{40}, \bibinfo{number}{4}, Article \bibinfo{articleno}{43} (\bibinfo{date}{jul} \bibinfo{year}{2021}), \bibinfo{numpages}{21}~pages.
\newblock
\urldef\tempurl%
\url{https://doi.org/10.1145/3450626.3459872}
\showDOI{\tempurl}


\bibitem[Parmar et~al\mbox{.}(2023)]%
        {parmar2023zero}
\bibfield{author}{\bibinfo{person}{Gaurav Parmar}, \bibinfo{person}{Krishna Kumar~Singh}, \bibinfo{person}{Richard Zhang}, \bibinfo{person}{Yijun Li}, \bibinfo{person}{Jingwan Lu}, {and} \bibinfo{person}{Jun-Yan Zhu}.} \bibinfo{year}{2023}\natexlab{}.
\newblock \showarticletitle{Zero-shot image-to-image translation}. In \bibinfo{booktitle}{\emph{ACM SIGGRAPH 2023 Conference Proceedings}}. \bibinfo{pages}{1--11}.
\newblock


\bibitem[Peers et~al\mbox{.}(2007)]%
        {peers2007_postproduction_facial}
\bibfield{author}{\bibinfo{person}{Pieter Peers}, \bibinfo{person}{Naoki Tamura}, \bibinfo{person}{Wojciech Matusik}, {and} \bibinfo{person}{Paul~E. Debevec}.} \bibinfo{year}{2007}\natexlab{}.
\newblock \showarticletitle{Post-production facial performance relighting using reflectance transfer}.
\newblock \bibinfo{journal}{\emph{{ACM} Trans. Graph.}} \bibinfo{volume}{26}, \bibinfo{number}{3} (\bibinfo{year}{2007}), \bibinfo{pages}{52}.
\newblock
\urldef\tempurl%
\url{https://doi.org/10.1145/1276377.1276442}
\showDOI{\tempurl}


\bibitem[Ponglertnapakorn et~al\mbox{.}(2023)]%
        {ponglertnapakorn2023difareli}
\bibfield{author}{\bibinfo{person}{Puntawat Ponglertnapakorn}, \bibinfo{person}{Nontawat Tritrong}, {and} \bibinfo{person}{Supasorn Suwajanakorn}.} \bibinfo{year}{2023}\natexlab{}.
\newblock \showarticletitle{DiFaReli: Diffusion face relighting}. In \bibinfo{booktitle}{\emph{Proceedings of the IEEE/CVF International Conference on Computer Vision}}. \bibinfo{pages}{22646--22657}.
\newblock


\bibitem[Preechakul et~al\mbox{.}(2022)]%
        {preechakul2022diffusion}
\bibfield{author}{\bibinfo{person}{Konpat Preechakul}, \bibinfo{person}{Nattanat Chatthee}, \bibinfo{person}{Suttisak Wizadwongsa}, {and} \bibinfo{person}{Supasorn Suwajanakorn}.} \bibinfo{year}{2022}\natexlab{}.
\newblock \showarticletitle{Diffusion autoencoders: Toward a meaningful and decodable representation}. In \bibinfo{booktitle}{\emph{Proceedings of the IEEE/CVF Conference on Computer Vision and Pattern Recognition}}. \bibinfo{pages}{10619--10629}.
\newblock


\bibitem[Rao et~al\mbox{.}(2022)]%
        {rao2022vorf}
\bibfield{author}{\bibinfo{person}{Pramod Rao}, \bibinfo{person}{Mallikarjun B~R}, \bibinfo{person}{Gereon Fox}, \bibinfo{person}{Tim Weyrich}, \bibinfo{person}{Bernd Bickel}, \bibinfo{person}{Hans-Peter Seidel}, \bibinfo{person}{Hanspeter Pfister}, \bibinfo{person}{Wojciech Matusik}, \bibinfo{person}{Ayush Tewari}, \bibinfo{person}{Christian Theobalt}, {and} \bibinfo{person}{Mohamed Elgharib}.} \bibinfo{year}{2022}\natexlab{}.
\newblock \showarticletitle{VoRF: Volumetric Relightable Faces}.
\newblock \bibinfo{journal}{\emph{BMVC}} (\bibinfo{year}{2022}).
\newblock


\bibitem[Rombach et~al\mbox{.}(2022)]%
        {rombach2022high}
\bibfield{author}{\bibinfo{person}{Robin Rombach}, \bibinfo{person}{Andreas Blattmann}, \bibinfo{person}{Dominik Lorenz}, \bibinfo{person}{Patrick Esser}, {and} \bibinfo{person}{Bj{\"o}rn Ommer}.} \bibinfo{year}{2022}\natexlab{}.
\newblock \showarticletitle{High-resolution image synthesis with latent diffusion models}. In \bibinfo{booktitle}{\emph{Proceedings of the IEEE/CVF conference on computer vision and pattern recognition}}. \bibinfo{pages}{10684--10695}.
\newblock


\bibitem[Ronneberger et~al\mbox{.}(2015)]%
        {Ronneberger2015_unet}
\bibfield{author}{\bibinfo{person}{Olaf Ronneberger}, \bibinfo{person}{Philipp Fischer}, {and} \bibinfo{person}{Thomas Brox}.} \bibinfo{year}{2015}\natexlab{}.
\newblock \showarticletitle{U-Net: Convolutional Networks for Biomedical Image Segmentation}. In \bibinfo{booktitle}{\emph{Medical Image Computing and Computer-Assisted Intervention -- MICCAI 2015}}, \bibfield{editor}{\bibinfo{person}{Nassir Navab}, \bibinfo{person}{Joachim Hornegger}, \bibinfo{person}{William~M. Wells}, {and} \bibinfo{person}{Alejandro~F. Frangi}} (Eds.). \bibinfo{publisher}{Springer International Publishing}, \bibinfo{address}{Cham}, \bibinfo{pages}{234--241}.
\newblock
\showISBNx{978-3-319-24574-4}


\bibitem[Saito et~al\mbox{.}(2023)]%
        {saito2023_relightable_gs_codec_avatars}
\bibfield{author}{\bibinfo{person}{Shunsuke Saito}, \bibinfo{person}{Tomas Simon}, \bibinfo{person}{Junxuan Li}, {and} \bibinfo{person}{Giljoo Nam}.} \bibinfo{year}{2023}\natexlab{}.
\newblock \showarticletitle{Relightable Gaussian Codec Avatars}.
\newblock \bibinfo{journal}{\emph{CoRR}}  \bibinfo{volume}{abs/2312.03704} (\bibinfo{year}{2023}).
\newblock
\urldef\tempurl%
\url{https://doi.org/10.48550/ARXIV.2312.03704}
\showDOI{\tempurl}
\showeprint[arXiv]{2312.03704}


\bibitem[Sarkar et~al\mbox{.}(2023)]%
        {sarkar2023_litnerf}
\bibfield{author}{\bibinfo{person}{Kripasindhu Sarkar}, \bibinfo{person}{Marcel~C. B{\"{u}}hler}, \bibinfo{person}{Gengyan Li}, \bibinfo{person}{Daoye Wang}, \bibinfo{person}{Delio Vicini}, \bibinfo{person}{J{\'{e}}r{\'{e}}my Riviere}, \bibinfo{person}{Yinda Zhang}, \bibinfo{person}{Sergio Orts{-}Escolano}, \bibinfo{person}{Paulo F.~U. Gotardo}, \bibinfo{person}{Thabo Beeler}, {and} \bibinfo{person}{Abhimitra Meka}.} \bibinfo{year}{2023}\natexlab{}.
\newblock \showarticletitle{LitNeRF: Intrinsic Radiance Decomposition for High-Quality View Synthesis and Relighting of Faces}. In \bibinfo{booktitle}{\emph{{SIGGRAPH} Asia 2023 Conference Papers, {SA} 2023, Sydney, NSW, Australia, December 12-15, 2023}} (<conf-loc>, <city>Sydney</city>, <state>NSW</state>, <country>Australia</country>, </conf-loc>) \emph{(\bibinfo{series}{SA '23})}, \bibfield{editor}{\bibinfo{person}{June Kim}, \bibinfo{person}{Ming~C. Lin}, {and} \bibinfo{person}{Bernd Bickel}} (Eds.). \bibinfo{publisher}{{ACM}}, Article \bibinfo{articleno}{42}, \bibinfo{numpages}{11}~pages.
\newblock
\urldef\tempurl%
\url{https://doi.org/10.1145/3610548.3618210}
\showDOI{\tempurl}


\bibitem[Sato et~al\mbox{.}(1997)]%
        {Sato:1997:Reflectance}
\bibfield{author}{\bibinfo{person}{Yoichi Sato}, \bibinfo{person}{Mark~D. Wheeler}, {and} \bibinfo{person}{Katsushi Ikeuchi}.} \bibinfo{year}{1997}\natexlab{}.
\newblock \showarticletitle{Object shape and reflectance modeling from observation}. In \bibinfo{booktitle}{\emph{Proceedings of the 24th Annual Conference on Computer Graphics and Interactive Techniques}} \emph{(\bibinfo{series}{SIGGRAPH '97})}. \bibinfo{publisher}{ACM Press/Addison-Wesley Publishing Co.}, \bibinfo{address}{USA}, \bibinfo{pages}{379–387}.
\newblock
\showISBNx{0897918967}
\urldef\tempurl%
\url{https://doi.org/10.1145/258734.258885}
\showDOI{\tempurl}


\bibitem[Schuhmann et~al\mbox{.}(2022)]%
        {clip}
\bibfield{author}{\bibinfo{person}{Christoph Schuhmann}, \bibinfo{person}{Romain Beaumont}, \bibinfo{person}{Richard Vencu}, \bibinfo{person}{Cade~W Gordon}, \bibinfo{person}{Ross Wightman}, \bibinfo{person}{Mehdi Cherti}, \bibinfo{person}{Theo Coombes}, \bibinfo{person}{Aarush Katta}, \bibinfo{person}{Clayton Mullis}, \bibinfo{person}{Mitchell Wortsman}, \bibinfo{person}{Patrick Schramowski}, \bibinfo{person}{Srivatsa~R Kundurthy}, \bibinfo{person}{Katherine Crowson}, \bibinfo{person}{Ludwig Schmidt}, \bibinfo{person}{Robert Kaczmarczyk}, {and} \bibinfo{person}{Jenia Jitsev}.} \bibinfo{year}{2022}\natexlab{}.
\newblock \showarticletitle{{LAION}-5B: An open large-scale dataset for training next generation image-text models}. In \bibinfo{booktitle}{\emph{Thirty-sixth Conference on Neural Information Processing Systems Datasets and Benchmarks Track}}.
\newblock
\urldef\tempurl%
\url{https://openreview.net/forum?id=M3Y74vmsMcY}
\showURL{%
\tempurl}


\bibitem[Sengupta et~al\mbox{.}(2018)]%
        {sengupta2018sfsnet}
\bibfield{author}{\bibinfo{person}{Soumyadip Sengupta}, \bibinfo{person}{Angjoo Kanazawa}, \bibinfo{person}{Carlos~D Castillo}, {and} \bibinfo{person}{David~W Jacobs}.} \bibinfo{year}{2018}\natexlab{}.
\newblock \showarticletitle{Sfsnet: Learning shape, reflectance and illuminance of facesin the wild'}. In \bibinfo{booktitle}{\emph{Proceedings of the IEEE conference on computer vision and pattern recognition}}. \bibinfo{pages}{6296--6305}.
\newblock


\bibitem[Shih et~al\mbox{.}(2014)]%
        {shih2014style}
\bibfield{author}{\bibinfo{person}{YiChang Shih}, \bibinfo{person}{Sylvain Paris}, \bibinfo{person}{Connelly Barnes}, \bibinfo{person}{William~T Freeman}, {and} \bibinfo{person}{Fr{\'e}do Durand}.} \bibinfo{year}{2014}\natexlab{}.
\newblock \showarticletitle{Style transfer for headshot portraits}.
\newblock  (\bibinfo{year}{2014}).
\newblock


\bibitem[Shu et~al\mbox{.}(2017a)]%
        {shu2017portrait}
\bibfield{author}{\bibinfo{person}{Zhixin Shu}, \bibinfo{person}{Sunil Hadap}, \bibinfo{person}{Eli Shechtman}, \bibinfo{person}{Kalyan Sunkavalli}, \bibinfo{person}{Sylvain Paris}, {and} \bibinfo{person}{Dimitris Samaras}.} \bibinfo{year}{2017}\natexlab{a}.
\newblock \showarticletitle{Portrait lighting transfer using a mass transport approach}.
\newblock \bibinfo{journal}{\emph{ACM Transactions on Graphics (TOG)}} \bibinfo{volume}{36}, \bibinfo{number}{4} (\bibinfo{year}{2017}), \bibinfo{pages}{1}.
\newblock


\bibitem[Shu et~al\mbox{.}(2017b)]%
        {shu2017neural}
\bibfield{author}{\bibinfo{person}{Zhixin Shu}, \bibinfo{person}{Ersin Yumer}, \bibinfo{person}{Sunil Hadap}, \bibinfo{person}{Kalyan Sunkavalli}, \bibinfo{person}{Eli Shechtman}, {and} \bibinfo{person}{Dimitris Samaras}.} \bibinfo{year}{2017}\natexlab{b}.
\newblock \showarticletitle{Neural face editing with intrinsic image disentangling}. In \bibinfo{booktitle}{\emph{Proceedings of the IEEE conference on computer vision and pattern recognition}}. \bibinfo{pages}{5541--5550}.
\newblock


\bibitem[Song et~al\mbox{.}(2020a)]%
        {song2020denoising}
\bibfield{author}{\bibinfo{person}{Jiaming Song}, \bibinfo{person}{Chenlin Meng}, {and} \bibinfo{person}{Stefano Ermon}.} \bibinfo{year}{2020}\natexlab{a}.
\newblock \showarticletitle{Denoising diffusion implicit models}.
\newblock \bibinfo{journal}{\emph{arXiv preprint arXiv:2010.02502}} (\bibinfo{year}{2020}).
\newblock


\bibitem[Song et~al\mbox{.}(2020b)]%
        {song2020score}
\bibfield{author}{\bibinfo{person}{Yang Song}, \bibinfo{person}{Jascha Sohl-Dickstein}, \bibinfo{person}{Diederik~P Kingma}, \bibinfo{person}{Abhishek Kumar}, \bibinfo{person}{Stefano Ermon}, {and} \bibinfo{person}{Ben Poole}.} \bibinfo{year}{2020}\natexlab{b}.
\newblock \showarticletitle{Score-based generative modeling through stochastic differential equations}.
\newblock \bibinfo{journal}{\emph{arXiv preprint arXiv:2011.13456}} (\bibinfo{year}{2020}).
\newblock


\bibitem[Srinivasan et~al\mbox{.}(2021)]%
        {srinivasan2021_nerv}
\bibfield{author}{\bibinfo{person}{Pratul~P. Srinivasan}, \bibinfo{person}{Boyang Deng}, \bibinfo{person}{Xiuming Zhang}, \bibinfo{person}{Matthew Tancik}, \bibinfo{person}{Ben Mildenhall}, {and} \bibinfo{person}{Jonathan~T. Barron}.} \bibinfo{year}{2021}\natexlab{}.
\newblock \showarticletitle{NeRV: Neural Reflectance and Visibility Fields for Relighting and View Synthesis}. In \bibinfo{booktitle}{\emph{{IEEE} Conference on Computer Vision and Pattern Recognition, {CVPR} 2021, virtual, June 19-25, 2021}}. \bibinfo{publisher}{Computer Vision Foundation / {IEEE}}, \bibinfo{pages}{7495--7504}.
\newblock
\urldef\tempurl%
\url{https://doi.org/10.1109/CVPR46437.2021.00741}
\showDOI{\tempurl}


\bibitem[Sun et~al\mbox{.}(2019)]%
        {sun2019single}
\bibfield{author}{\bibinfo{person}{Tiancheng Sun}, \bibinfo{person}{Jonathan~T Barron}, \bibinfo{person}{Yun-Ta Tsai}, \bibinfo{person}{Zexiang Xu}, \bibinfo{person}{Xueming Yu}, \bibinfo{person}{Graham Fyffe}, \bibinfo{person}{Christoph Rhemann}, \bibinfo{person}{Jay Busch}, \bibinfo{person}{Paul Debevec}, {and} \bibinfo{person}{Ravi Ramamoorthi}.} \bibinfo{year}{2019}\natexlab{}.
\newblock \showarticletitle{Single image portrait relighting}.
\newblock \bibinfo{journal}{\emph{ACM Transactions on Graphics (TOG)}} \bibinfo{volume}{38}, \bibinfo{number}{4} (\bibinfo{year}{2019}), \bibinfo{pages}{1--12}.
\newblock


\bibitem[Tancik et~al\mbox{.}(2023)]%
        {tancik2023nerfstudio}
\bibfield{author}{\bibinfo{person}{Matthew Tancik}, \bibinfo{person}{Ethan Weber}, \bibinfo{person}{Evonne Ng}, \bibinfo{person}{Ruilong Li}, \bibinfo{person}{Brent Yi}, \bibinfo{person}{Terrance Wang}, \bibinfo{person}{Alexander Kristoffersen}, \bibinfo{person}{Jake Austin}, \bibinfo{person}{Kamyar Salahi}, \bibinfo{person}{Abhik Ahuja}, {et~al\mbox{.}}} \bibinfo{year}{2023}\natexlab{}.
\newblock \showarticletitle{Nerfstudio: A modular framework for neural radiance field development}. In \bibinfo{booktitle}{\emph{ACM SIGGRAPH 2023 Conference Proceedings}}. \bibinfo{pages}{1--12}.
\newblock


\bibitem[Tewari et~al\mbox{.}(2021a)]%
        {tewari2021photoapp}
\bibfield{author}{\bibinfo{person}{Ayush Tewari}, \bibinfo{person}{Abdallah Dib}, \bibinfo{person}{Tim Weyrich}, \bibinfo{person}{Bernd Bickel}, \bibinfo{person}{Hans-Peter Seidel}, \bibinfo{person}{Hanspeter Pfister}, \bibinfo{person}{Wojciech Matusik}, \bibinfo{person}{Louis Chevallier}, \bibinfo{person}{Mohamed Elgharib}, \bibinfo{person}{Christian Theobalt}, {et~al\mbox{.}}} \bibinfo{year}{2021}\natexlab{a}.
\newblock \showarticletitle{PhotoApp: Photorealistic Appearance Editing of Head Portraits}.
\newblock \bibinfo{journal}{\emph{arXiv preprint arXiv:2103.07658}} (\bibinfo{year}{2021}).
\newblock


\bibitem[Tewari et~al\mbox{.}(2020)]%
        {tewari2020pie}
\bibfield{author}{\bibinfo{person}{Ayush Tewari}, \bibinfo{person}{Mohamed Elgharib}, \bibinfo{person}{Florian Bernard}, \bibinfo{person}{Hans-Peter Seidel}, \bibinfo{person}{Patrick P{\'e}rez}, \bibinfo{person}{Michael Zollh{\"o}fer}, {and} \bibinfo{person}{Christian Theobalt}.} \bibinfo{year}{2020}\natexlab{}.
\newblock \showarticletitle{Pie: Portrait image embedding for semantic control}.
\newblock \bibinfo{journal}{\emph{ACM Transactions on Graphics (TOG)}} \bibinfo{volume}{39}, \bibinfo{number}{6} (\bibinfo{year}{2020}), \bibinfo{pages}{1--14}.
\newblock


\bibitem[Tewari et~al\mbox{.}(2021b)]%
        {tewari2021monocular}
\bibfield{author}{\bibinfo{person}{Ayush Tewari}, \bibinfo{person}{Tae-Hyun Oh}, \bibinfo{person}{Tim Weyrich}, \bibinfo{person}{Bernd Bickel}, \bibinfo{person}{Hans-Peter Seidel}, \bibinfo{person}{Hanspeter Pfister}, \bibinfo{person}{Wojciech Matusik}, \bibinfo{person}{Mohamed Elgharib}, \bibinfo{person}{Christian Theobalt}, {et~al\mbox{.}}} \bibinfo{year}{2021}\natexlab{b}.
\newblock \showarticletitle{Monocular reconstruction of neural face reflectance fields}. In \bibinfo{booktitle}{\emph{Proceedings of the IEEE/CVF Conference on Computer Vision and Pattern Recognition}}. \bibinfo{pages}{4791--4800}.
\newblock


\bibitem[Wang et~al\mbox{.}(2023b)]%
        {wang2023exploiting}
\bibfield{author}{\bibinfo{person}{Jianyi Wang}, \bibinfo{person}{Zongsheng Yue}, \bibinfo{person}{Shangchen Zhou}, \bibinfo{person}{Kelvin~CK Chan}, {and} \bibinfo{person}{Chen~Change Loy}.} \bibinfo{year}{2023}\natexlab{b}.
\newblock \showarticletitle{Exploiting diffusion prior for real-world image super-resolution}.
\newblock \bibinfo{journal}{\emph{arXiv preprint arXiv:2305.07015}} (\bibinfo{year}{2023}).
\newblock


\bibitem[Wang et~al\mbox{.}(2023a)]%
        {wang2023_neus}
\bibfield{author}{\bibinfo{person}{Peng Wang}, \bibinfo{person}{Lingjie Liu}, \bibinfo{person}{Yuan Liu}, \bibinfo{person}{Christian Theobalt}, \bibinfo{person}{Taku Komura}, {and} \bibinfo{person}{Wenping Wang}.} \bibinfo{year}{2023}\natexlab{a}.
\newblock \bibinfo{title}{NeuS: Learning Neural Implicit Surfaces by Volume Rendering for Multi-view Reconstruction}.
\newblock
\newblock
\showeprint[arxiv]{2106.10689}~[cs.CV]


\bibitem[Wang et~al\mbox{.}(2008)]%
        {wang2008face}
\bibfield{author}{\bibinfo{person}{Yang Wang}, \bibinfo{person}{Lei Zhang}, \bibinfo{person}{Zicheng Liu}, \bibinfo{person}{Gang Hua}, \bibinfo{person}{Zhen Wen}, \bibinfo{person}{Zhengyou Zhang}, {and} \bibinfo{person}{Dimitris Samaras}.} \bibinfo{year}{2008}\natexlab{}.
\newblock \showarticletitle{Face relighting from a single image under arbitrary unknown lighting conditions}.
\newblock \bibinfo{journal}{\emph{IEEE transactions on pattern analysis and machine intelligence}} \bibinfo{volume}{31}, \bibinfo{number}{11} (\bibinfo{year}{2008}), \bibinfo{pages}{1968--1984}.
\newblock


\bibitem[Weyrich et~al\mbox{.}(2006)]%
        {weyrich2006_analysis_of_human_faces}
\bibfield{author}{\bibinfo{person}{Tim Weyrich}, \bibinfo{person}{Wojciech Matusik}, \bibinfo{person}{Hanspeter Pfister}, \bibinfo{person}{Bernd Bickel}, \bibinfo{person}{Craig Donner}, \bibinfo{person}{Chien Tu}, \bibinfo{person}{Janet McAndless}, \bibinfo{person}{Jinho Lee}, \bibinfo{person}{Addy Ngan}, \bibinfo{person}{Henrik~Wann Jensen}, {and} \bibinfo{person}{Markus~H. Gross}.} \bibinfo{year}{2006}\natexlab{}.
\newblock \showarticletitle{Analysis of human faces using a measurement-based skin reflectance model}.
\newblock \bibinfo{journal}{\emph{{ACM} Trans. Graph.}} \bibinfo{volume}{25}, \bibinfo{number}{3} (\bibinfo{year}{2006}), \bibinfo{pages}{1013--1024}.
\newblock
\urldef\tempurl%
\url{https://doi.org/10.1145/1141911.1141987}
\showDOI{\tempurl}


\bibitem[Whitaker(2024)]%
        {multiresnoise}
\bibfield{author}{\bibinfo{person}{John Whitaker}.} \bibinfo{year}{2024}\natexlab{}.
\newblock \bibinfo{title}{Multi-Resolution Noise for Diffusion Model Training}.
\newblock \bibinfo{howpublished}{\url{https://wandb.ai/johnowhitaker/multires_noise/reports/Multi-Resolution-Noise-for-Diffusion-Model-Training--VmlldzozNjYyOTU2}}.
\newblock
\newblock
\shownote{Accessed: 2024-05-15}.


\bibitem[Wu et~al\mbox{.}(2023)]%
        {wu2023_4d_gs}
\bibfield{author}{\bibinfo{person}{Guanjun Wu}, \bibinfo{person}{Taoran Yi}, \bibinfo{person}{Jiemin Fang}, \bibinfo{person}{Lingxi Xie}, \bibinfo{person}{Xiaopeng Zhang}, \bibinfo{person}{Wei Wei}, \bibinfo{person}{Wenyu Liu}, \bibinfo{person}{Qi Tian}, {and} \bibinfo{person}{Xinggang Wang}.} \bibinfo{year}{2023}\natexlab{}.
\newblock \showarticletitle{4D Gaussian Splatting for Real-Time Dynamic Scene Rendering}.
\newblock \bibinfo{journal}{\emph{CoRR}}  \bibinfo{volume}{abs/2310.08528} (\bibinfo{year}{2023}).
\newblock
\urldef\tempurl%
\url{https://doi.org/10.48550/ARXIV.2310.08528}
\showDOI{\tempurl}
\showeprint[arXiv]{2310.08528}


\bibitem[Xu et~al\mbox{.}(2023a)]%
        {xu2023artistfriendlyrelightableanimatableneural}
\bibfield{author}{\bibinfo{person}{Yingyan Xu}, \bibinfo{person}{Prashanth Chandran}, \bibinfo{person}{Sebastian Weiss}, \bibinfo{person}{Markus Gross}, \bibinfo{person}{Gaspard Zoss}, {and} \bibinfo{person}{Derek Bradley}.} \bibinfo{year}{2023}\natexlab{a}.
\newblock \bibinfo{title}{Artist-Friendly Relightable and Animatable Neural Heads}.
\newblock
\newblock
\showeprint[arxiv]{2312.03420}~[cs.CV]
\urldef\tempurl%
\url{https://arxiv.org/abs/2312.03420}
\showURL{%
\tempurl}


\bibitem[Xu et~al\mbox{.}(2023b)]%
        {xu2023_renerf}
\bibfield{author}{\bibinfo{person}{Yingyan Xu}, \bibinfo{person}{Gaspard Zoss}, \bibinfo{person}{Prashanth Chandran}, \bibinfo{person}{Markus Gross}, \bibinfo{person}{Derek Bradley}, {and} \bibinfo{person}{Paulo Gotardo}.} \bibinfo{year}{2023}\natexlab{b}.
\newblock \showarticletitle{ReNeRF: Relightable Neural Radiance Fields with Nearfield Lighting}. In \bibinfo{booktitle}{\emph{International Conference on Computer Vision (ICCV)}}.
\newblock


\bibitem[Yang et~al\mbox{.}(2023)]%
        {yang2023practicalcapturehighfidelityrelightable}
\bibfield{author}{\bibinfo{person}{Haotian Yang}, \bibinfo{person}{Mingwu Zheng}, \bibinfo{person}{Wanquan Feng}, \bibinfo{person}{Haibin Huang}, \bibinfo{person}{Yu-Kun Lai}, \bibinfo{person}{Pengfei Wan}, \bibinfo{person}{Zhongyuan Wang}, {and} \bibinfo{person}{Chongyang Ma}.} \bibinfo{year}{2023}\natexlab{}.
\newblock \bibinfo{title}{Towards Practical Capture of High-Fidelity Relightable Avatars}.
\newblock
\newblock
\showeprint[arxiv]{2309.04247}~[cs.CV]
\urldef\tempurl%
\url{https://arxiv.org/abs/2309.04247}
\showURL{%
\tempurl}


\bibitem[Yao et~al\mbox{.}(2022)]%
        {yao2022_neilf}
\bibfield{author}{\bibinfo{person}{Yao Yao}, \bibinfo{person}{Jingyang Zhang}, \bibinfo{person}{Jingbo Liu}, \bibinfo{person}{Yihang Qu}, \bibinfo{person}{Tian Fang}, \bibinfo{person}{David McKinnon}, \bibinfo{person}{Yanghai Tsin}, {and} \bibinfo{person}{Long Quan}.} \bibinfo{year}{2022}\natexlab{}.
\newblock \showarticletitle{NeILF: Neural Incident Light Field for Physically-based Material Estimation}. In \bibinfo{booktitle}{\emph{Computer Vision - {ECCV} 2022 - 17th European Conference, Tel Aviv, Israel, October 23-27, 2022, Proceedings, Part {XXXI}}} \emph{(\bibinfo{series}{Lecture Notes in Computer Science}, Vol.~\bibinfo{volume}{13691})}, \bibfield{editor}{\bibinfo{person}{Shai Avidan}, \bibinfo{person}{Gabriel~J. Brostow}, \bibinfo{person}{Moustapha Ciss{\'{e}}}, \bibinfo{person}{Giovanni~Maria Farinella}, {and} \bibinfo{person}{Tal Hassner}} (Eds.). \bibinfo{publisher}{Springer}, \bibinfo{pages}{700--716}.
\newblock
\urldef\tempurl%
\url{https://doi.org/10.1007/978-3-031-19821-2\_40}
\showDOI{\tempurl}


\bibitem[Zeng et~al\mbox{.}(2023)]%
        {zeng2023_relighting_nerf_shadow}
\bibfield{author}{\bibinfo{person}{Chong Zeng}, \bibinfo{person}{Guojun Chen}, \bibinfo{person}{Yue Dong}, \bibinfo{person}{Pieter Peers}, \bibinfo{person}{Hongzhi Wu}, {and} \bibinfo{person}{Xin Tong}.} \bibinfo{year}{2023}\natexlab{}.
\newblock \showarticletitle{Relighting Neural Radiance Fields with Shadow and Highlight Hints}. In \bibinfo{booktitle}{\emph{ACM SIGGRAPH 2023 Conference Proceedings}} (Los Angeles, CA, USA) \emph{(\bibinfo{series}{SIGGRAPH '23})}. \bibinfo{publisher}{Association for Computing Machinery}, \bibinfo{address}{New York, NY, USA}, Article \bibinfo{articleno}{73}, \bibinfo{numpages}{11}~pages.
\newblock
\showISBNx{9798400701597}
\urldef\tempurl%
\url{https://doi.org/10.1145/3588432.3591482}
\showDOI{\tempurl}


\bibitem[Zeng et~al\mbox{.}(2024)]%
        {zeng2024dilightnet}
\bibfield{author}{\bibinfo{person}{Chong Zeng}, \bibinfo{person}{Yue Dong}, \bibinfo{person}{Pieter Peers}, \bibinfo{person}{Youkang Kong}, \bibinfo{person}{Hongzhi Wu}, {and} \bibinfo{person}{Xin Tong}.} \bibinfo{year}{2024}\natexlab{}.
\newblock \showarticletitle{DiLightNet: Fine-grained Lighting Control for Diffusion-based Image Generation}.
\newblock \bibinfo{journal}{\emph{arXiv preprint arXiv:2402.11929}} (\bibinfo{year}{2024}).
\newblock


\bibitem[Zhang et~al\mbox{.}(2023b)]%
        {zhang2023_neilfpp}
\bibfield{author}{\bibinfo{person}{Jingyang Zhang}, \bibinfo{person}{Yao Yao}, \bibinfo{person}{Shiwei Li}, \bibinfo{person}{Jingbo Liu}, \bibinfo{person}{Tian Fang}, \bibinfo{person}{David McKinnon}, \bibinfo{person}{Yanghai Tsin}, {and} \bibinfo{person}{Long Quan}.} \bibinfo{year}{2023}\natexlab{b}.
\newblock \showarticletitle{NeILF++: Inter-Reflectable Light Fields for Geometry and Material Estimation}.
\newblock \bibinfo{journal}{\emph{CoRR}}  \bibinfo{volume}{abs/2303.17147} (\bibinfo{year}{2023}).
\newblock
\urldef\tempurl%
\url{https://doi.org/10.48550/ARXIV.2303.17147}
\showDOI{\tempurl}
\showeprint[arXiv]{2303.17147}


\bibitem[Zhang et~al\mbox{.}(2023a)]%
        {zhang2023adding}
\bibfield{author}{\bibinfo{person}{Lvmin Zhang}, \bibinfo{person}{Anyi Rao}, {and} \bibinfo{person}{Maneesh Agrawala}.} \bibinfo{year}{2023}\natexlab{a}.
\newblock \showarticletitle{Adding conditional control to text-to-image diffusion models}. In \bibinfo{booktitle}{\emph{Proceedings of the IEEE/CVF International Conference on Computer Vision}}. \bibinfo{pages}{3836--3847}.
\newblock


\bibitem[Zhang et~al\mbox{.}(2024)]%
        {iclight}
\bibfield{author}{\bibinfo{person}{Lvmin Zhang}, \bibinfo{person}{Anyi Rao}, {and} \bibinfo{person}{Maneesh Agrawala}.} \bibinfo{year}{2024}\natexlab{}.
\newblock \bibinfo{title}{IC-Light GitHub Page}.
\newblock
\newblock


\bibitem[Zhang et~al\mbox{.}(2018)]%
        {zhang2018unreasonable}
\bibfield{author}{\bibinfo{person}{Richard Zhang}, \bibinfo{person}{Phillip Isola}, \bibinfo{person}{Alexei~A Efros}, \bibinfo{person}{Eli Shechtman}, {and} \bibinfo{person}{Oliver Wang}.} \bibinfo{year}{2018}\natexlab{}.
\newblock \showarticletitle{The unreasonable effectiveness of deep features as a perceptual metric}. In \bibinfo{booktitle}{\emph{Proceedings of the IEEE conference on computer vision and pattern recognition}}. \bibinfo{pages}{586--595}.
\newblock


\bibitem[Zhang et~al\mbox{.}(2021a)]%
        {zhang_neural_light_transport_2021}
\bibfield{author}{\bibinfo{person}{Xiuming Zhang}, \bibinfo{person}{Sean Fanello}, \bibinfo{person}{Yun-Ta Tsai}, \bibinfo{person}{Tiancheng Sun}, \bibinfo{person}{Tianfan Xue}, \bibinfo{person}{Rohit Pandey}, \bibinfo{person}{Sergio Orts-Escolano}, \bibinfo{person}{Philip Davidson}, \bibinfo{person}{Christoph Rhemann}, \bibinfo{person}{Paul Debevec}, \bibinfo{person}{Jonathan~T. Barron}, \bibinfo{person}{Ravi Ramamoorthi}, {and} \bibinfo{person}{William~T. Freeman}.} \bibinfo{year}{2021}\natexlab{a}.
\newblock \showarticletitle{Neural Light Transport for Relighting and View Synthesis}.
\newblock \bibinfo{journal}{\emph{ACM Trans. Graph.}} \bibinfo{volume}{40}, \bibinfo{number}{1}, Article \bibinfo{articleno}{9} (\bibinfo{date}{jan} \bibinfo{year}{2021}), \bibinfo{numpages}{17}~pages.
\newblock
\showISSN{0730-0301}
\urldef\tempurl%
\url{https://doi.org/10.1145/3446328}
\showDOI{\tempurl}


\bibitem[Zhang et~al\mbox{.}(2021b)]%
        {zhang2021_nerfactor}
\bibfield{author}{\bibinfo{person}{Xiuming Zhang}, \bibinfo{person}{Pratul~P. Srinivasan}, \bibinfo{person}{Boyang Deng}, \bibinfo{person}{Paul Debevec}, \bibinfo{person}{William~T. Freeman}, {and} \bibinfo{person}{Jonathan~T. Barron}.} \bibinfo{year}{2021}\natexlab{b}.
\newblock \showarticletitle{NeRFactor: neural factorization of shape and reflectance under an unknown illumination}.
\newblock \bibinfo{journal}{\emph{ACM Transactions on Graphics}} \bibinfo{volume}{40}, \bibinfo{number}{6} (\bibinfo{date}{Dec.} \bibinfo{year}{2021}), \bibinfo{pages}{1–18}.
\newblock
\showISSN{1557-7368}
\urldef\tempurl%
\url{https://doi.org/10.1145/3478513.3480496}
\showDOI{\tempurl}


\bibitem[Zhou et~al\mbox{.}(2019)]%
        {zhou2019deep}
\bibfield{author}{\bibinfo{person}{Hao Zhou}, \bibinfo{person}{Sunil Hadap}, \bibinfo{person}{Kalyan Sunkavalli}, {and} \bibinfo{person}{David~W Jacobs}.} \bibinfo{year}{2019}\natexlab{}.
\newblock \showarticletitle{Deep single-image portrait relighting}. In \bibinfo{booktitle}{\emph{Proceedings of the IEEE/CVF international conference on computer vision}}. \bibinfo{pages}{7194--7202}.
\newblock


\bibitem[Zwicker et~al\mbox{.}(2002)]%
        {zwicker2002ewa}
\bibfield{author}{\bibinfo{person}{M. Zwicker}, \bibinfo{person}{H. Pfister}, \bibinfo{person}{J. van Baar}, {and} \bibinfo{person}{M. Gross}.} \bibinfo{year}{2002}\natexlab{}.
\newblock \showarticletitle{EWA Splatting}.
\newblock \bibinfo{journal}{\emph{IEEE Transactions on Visualization and Computer Graphics}} \bibinfo{volume}{8}, \bibinfo{number}{3} (\bibinfo{date}{07/2002-09/2002} \bibinfo{year}{2002}), \bibinfo{pages}{223--238}.
\newblock


\end{thebibliography}
